\PassOptionsToPackage{dvipsnames}{xcolor}
\documentclass[twocolumn]{ModernAcademic-EN}

\usepackage{lipsum}       % Placeholder text
\usepackage{siunitx}      % Units and numbers
\usepackage{amsfonts}     % Additional math fonts
\usepackage{mathtools}    % Enhanced math support
\usepackage{booktabs}     % Professional tables
\usepackage{listings}     % Code listings
\usepackage{algorithm}
\usepackage{algpseudocode}
\usepackage{tikz}
\usetikzlibrary{positioning,arrows.meta}

\usepackage{nicefrac}       % compact symbols for 1/2, etc.
\usepackage{microtype}      % microtypography

\usepackage{graphicx}
\usepackage{subcaption} % for subfigures

\usepackage{amsmath}
\usepackage{amsthm}
\usepackage{amssymb}

\usepackage{pifont}
\usepackage{ulem}

\usepackage{float}
\usepackage{algorithm}

\usepackage{multirow}
\usepackage{adjustbox}
\usepackage{amssymb}

\usepackage{caption}
\usepackage{makecell}
\usepackage{graphicx}
\usepackage{algpseudocode}
\usepackage{algorithmicx}

\newcommand{\zhaoyuan}[1]{#1}
\newcommand{\greenup}{\textcolor{ForestGreen}{\ensuremath{\uparrow}}}  % 森林绿
\newcommand{\reddown}{\textcolor{red}{\ensuremath{\downarrow}}}
\usetikzlibrary{positioning,arrows.meta}

\setthemecolor{1a2a6c}

\settitlepreskip{-10ex}
\settitlelogogap{3em}

\titlelogocontent{%
	\begin{minipage}{\textwidth}
		\begin{minipage}[c]{0.35\textwidth}
			\raggedright
			\includegraphics[height=1cm]{figs/ntu_ai4x_logo_merge_1.png}% 团队logo
		\end{minipage}%
		\hfill
		\begin{minipage}[c]{0.60\textwidth}
			\raggedleft
			\includegraphics[height=1cm]{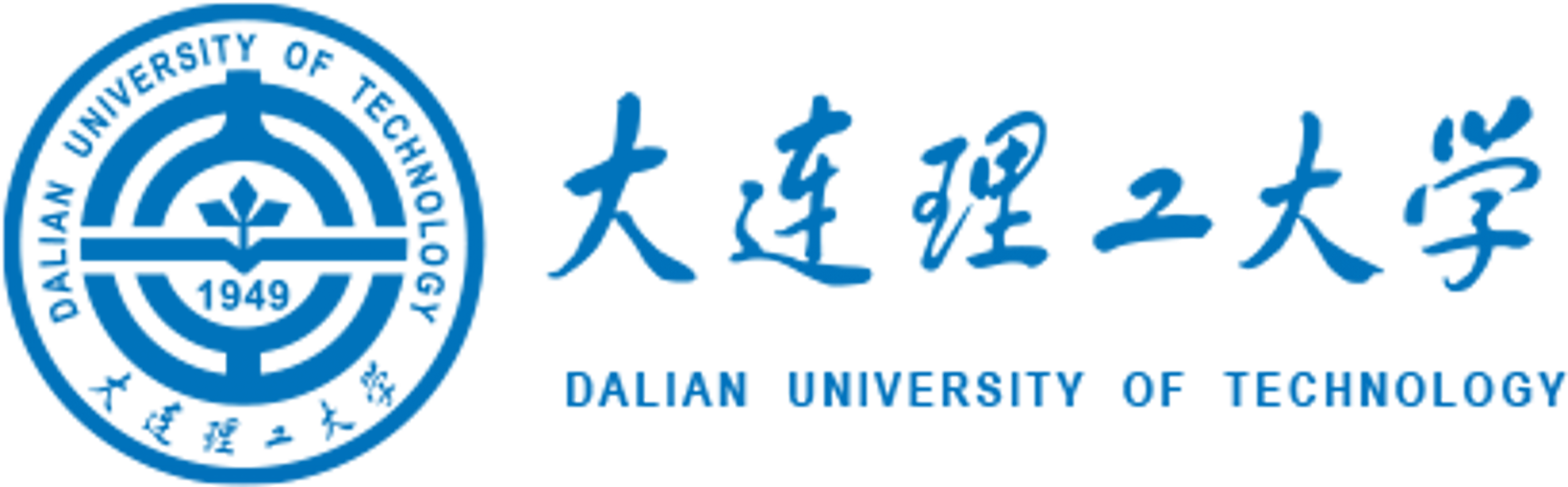}\hspace{0.5em}% 学校1
			\includegraphics[height=1cm]{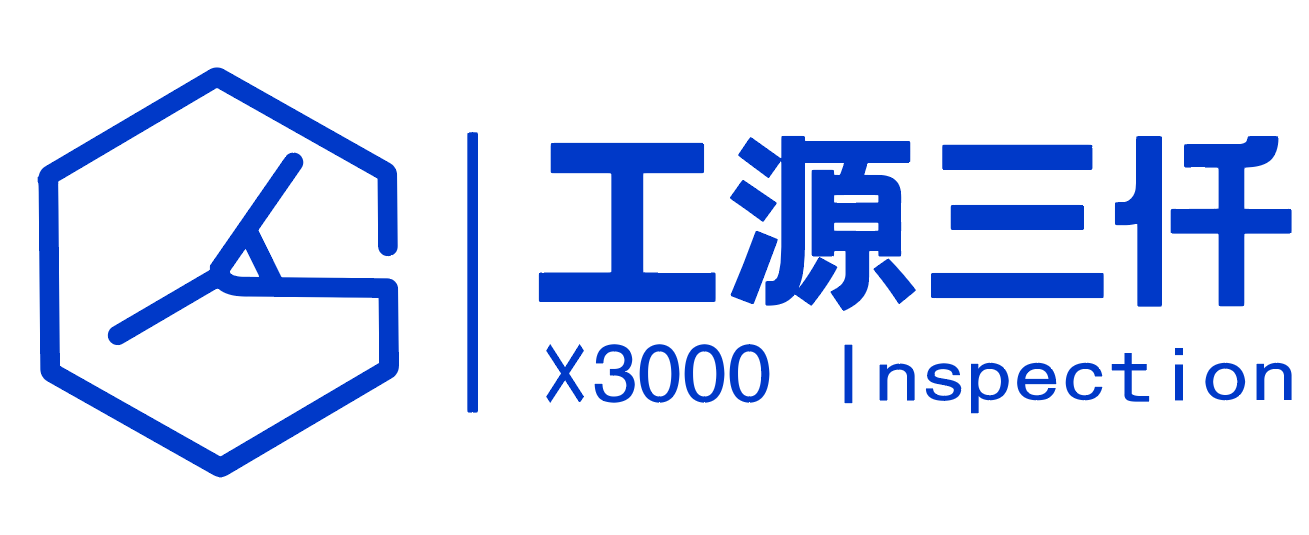}\hspace{0.5em}% 学校2
		\end{minipage}
		\\[0.4em]
		\textcolor{academicblue}{\rule{\textwidth}{0.6pt}}%
	\end{minipage}%
}

\showlogointitle
\addsociallink{\faGithub}{Code}{https://github.com/yuanzhao-CVLAB/UniMMAD}

\author{Yuan Zhao\textsuperscript{1,2*}, 
    Youwei Pang\textsuperscript{3*}, 
    Lihe Zhang\textsuperscript{1\#}, 
    Hanqi Liu\textsuperscript{2}, 
    \\
    Jiaming Zuo\textsuperscript{2\#}, 
    Huchuan Lu\textsuperscript{1},  
    Xiaoqi Zhao\textsuperscript{3\#}}
\shorttitle{UniMMAD}
\shortauthor{Y. Zhao et~al.}
\affiliation{
    \textsuperscript{1}IIAU-Lab, Dalian University of Technology,
    \textsuperscript{2}X3000 Inspection Co., Ltd,
    \textsuperscript{3}AI4X Team, Nanyang Technological University
}

\email{
\par
\textsuperscript{*}equal contribution. \textsuperscript{\#}corresponding authors.}

\title{UniMMAD: Unified Multi-Modal and Multi-Class Anomaly Detection via MoE-Driven Feature Decompression}
\abstract{
Existing anomaly detection  methods often treat  the  modality and class as independent factors. Although this paradigm has enriched the development of AD research branches and produced many specialized models, it has also led to fragmented solutions and excessive memory overhead. Moreover, reconstruction-based multi-class approaches typically rely on shared decoding paths, which struggle to handle large variations across domains, resulting in distorted normality boundaries, domain interference, and high false alarm rates.
To address these limitations, we propose UniMMAD, a unified framework for  multi-modal and multi-class anomaly detection. At the core of UniMMAD is a Mixture-of-Experts (MoE)-driven feature decompression mechanism, which enables adaptive and disentangled reconstruction tailored to specific domains.
This process is guided by a ``general → specific'' paradigm. 
In the encoding stage, multi-modal inputs of varying combinations are compressed into compact, general-purpose features. The encoder incorporates a feature compression module to suppress latent anomalies, encourage cross-modal interaction, and avoid shortcut learning.
In the decoding stage, the general features are decompressed into  modality-specific  and class-specific forms via a sparsely-gated cross MoE, which dynamically selects expert pathways based on input modality and class. To further improve efficiency, we design a grouped dynamic filtering mechanism and an MoE-in-MoE structure, reducing MoE parameter usage by approximately 75\% while maintaining sparse activation and fast inference.
 UniMMAD achieves state-of-the-art performance on 9 anomaly detection  datasets, spanning 3 fields, 12 modalities, and 66 classes.
}

\begin{document}

\maketitle

\section{Introduction}
Unsupervised anomaly detection (AD), which aims to detect anomalies without relying on abnormal samples,  is crucial in practical applications such as manufacturing inspection~\cite{bergmann2019mvtec} and medical diagnosis~\cite{li2018thoracic}. 
 In real-world scenarios, anomalies appear in diverse forms and often require multiple sensing modalities to be reliably detected. 
 For example, 3D modalities can help identify geometric defects and  mitigate the lighting sensitivity risks in  RGB images~\cite{costanzino2024multimodal}.

\begin{figure}[t] 
    \centering
    \includegraphics[width=1\linewidth]{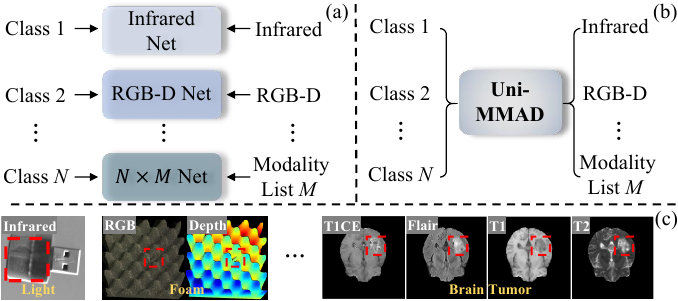}
    % \vspace{-5mm}
    \caption{ Task setting.
    (a) Existing methods~\cite{li2024multi,wang2023multimodal,costanzino2024multimodal} rely on specialized models tailored to the fixed modalities. (b) Our UniMMAD unifies multi-modal and multi-class anomaly detection within a single framework. (c) Visual examples, where anomaly regions, modalities and class names are marked by red boxes, white, and yellow, respectively. 
    }
    \label{fig:Teaser}
    \vspace{-6mm}
\end{figure}

\begin{figure*}[t] 
    \centering
    \includegraphics[width=0.96\linewidth]{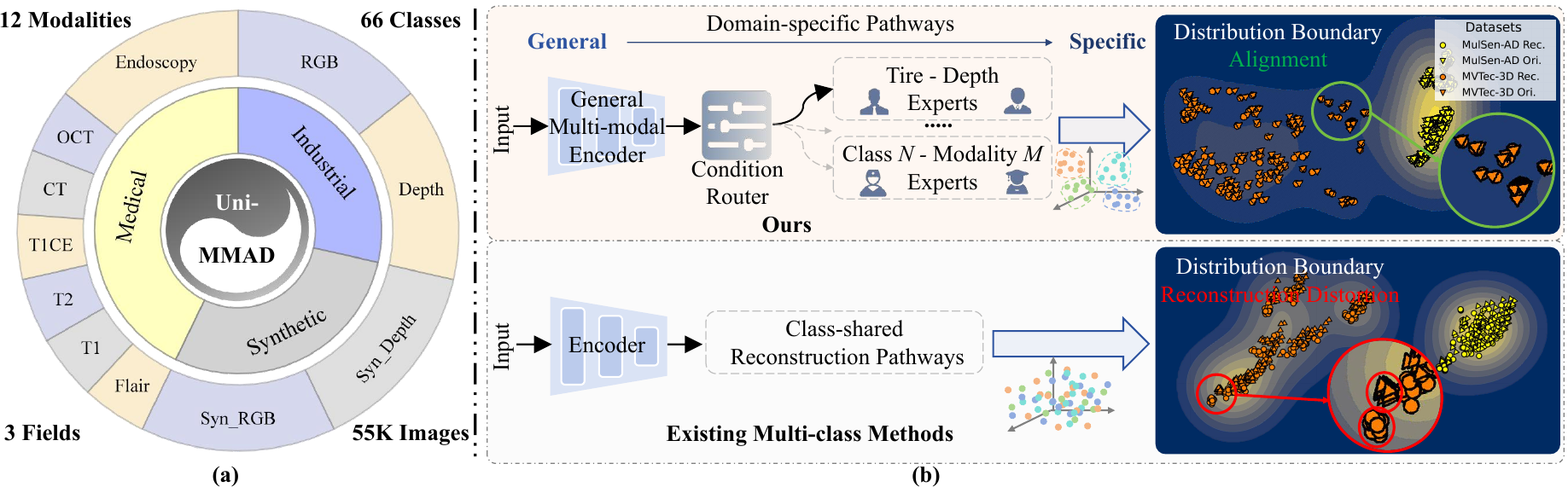}
    % \vspace{-6mm}
    \caption{
    (a) Overview of the fields, modalities, and classes encompassed by UniMMAD.
    (b) Architectures of UniMMAD and mainstream multi-class methods~\cite{you2022unified,he2024mambaad}. Reconstructed feature distribution in normal regions is shown on the right.
      Larger distances between original features (triangles $\blacktriangledown$) and reconstructed features (circles \ding{108}) mean a higher risk of false positives.
       }
    \label{fig:Teaser_Comparison}
    \vspace{-2mm}
\end{figure*}

Existing approaches~\cite{costanzino2024multimodal,wang2023multimodal,weninger2018segmentation}  primarily focus on 
the fixed modality inputs. They are dedicated to processing a kind of modality combination, as shown in Fig.~\ref{fig:Teaser}a. 
However, in real-world scenarios, multiple distinct sensors are typically involved for product quality inspection.
As shown in Fig.~\ref{fig:Teaser}c, %actual industrial manufacturing necessitates the use of 
infrared cameras are used to detect internal component damage in the ``Light'' product, while RGB and 3D sensors are essential for identifying color and geometric defects in the ``Foam'' product.
Customizing specific models for each modality combination is not an optimized solution and usually results in difficulties in model deployment and memory overhead. 
Although existing multi-class methods~\cite{you2022unified,luo2025exploring} can be extended to 
mitigate fragmentation issues with class-shared model designs, they struggle to adapt to complex modality combinations and diverse scenes, since shared decoders cannot disentangle heterogeneous modality distributions well.

Recently, unified vision models~\cite{zhao2024spider,wang2023seggpt,potlapalli2023promptir} have demonstrated that a single architecture and parameter set can handle various tasks. Inspired by these, we aim to develop a unified unsupervised  anomaly detection framework across multiple scenarios  that supports multi-modal inputs, and predicts multiple semantic classes, as illustrated in  Fig.~\ref{fig:Teaser}b  and Fig.~\ref{fig:Teaser_Comparison}a.
Achieving this goal poses two key challenges:
\textbf{1) Large domain heterogeneity.} The intricate interplay among diverse modalities and classes introduces large variations  in appearance, illumination, scale, background, and anomaly semantics, making consistent representation learning  and anomaly discrimination difficult.
\textbf{2) Practical constraints on efficiency and continual learning.} A practical unified AD model must ensure high accuracy, fast inference, sparse computation, and adapt to new classes or modalities without catastrophic  forgetting. Existing multi-class AD models~\cite{you2022unified,he2024mambaad} share a reconstruction path across classes. Although parameter-efficient, this shared  path aggravates domain interference, resulting in poor normal pattern reconstruction and high false alarms. As shown in Fig.~\ref{fig:Teaser_Comparison}b, these methods may distort decision boundaries and misclassify normal regions as anomalies.

To address these challenges, we propose UniMMAD, a unified anomaly detection framework for diverse scenes. Its core is a Mixture-of-Experts (MoE)-based feature decompression mechanism, which is designed to resolve the entanglement caused by heterogeneous inputs. Unified AD requires a shared model to handle large domain shifts across modalities and classes, which often leads to distorted reconstructions and false positives with a single decoder. MoE naturally alleviates this by enabling sparse, input-conditioned expert activation, letting experts specialize in domains while sharing global parameters. This allows decomposing general latent features into diverse domain-specific outputs. UniMMAD uses a unified encoder to compress arbitrary multi-modal inputs into compact features and a dynamic MoE decoder to adaptively decompress them into task-specific reconstructions with minimal interference. 

% %
As shown in Fig.~\ref{fig:Teaser_Comparison}b, we first introduce a novel paradigm, ``general → specific'', which decompresses  general multi-modal features into domain-specific  uni-modal ones. 
This paradigm encourages the learning of a powerful general-purpose multi-modal encoder that can flexibly adapt to diverse modality combinations and effectively fuse cross-modal information. 
% %
Unlike conventional reconstruction objectives~\cite{you2022unified} that  reconstruct the input itself, our asymmetric ``general → specific'' paradigm mitigates shortcut reconstruction. 
% %	
% %
Second, we embed a feature compression module in the encoder to generate  compact and   purified  latent representations. This compacts general features and curbs abnormal pattern propagation.
Third, in the decoding phase, we design Cross Mixture-of-Experts (C-MoE) to reduce domain interference during decompression. Inspired by the success of MoE on heterogeneous data~\cite{liu2024deepseek}, C-MoE uses a cross-conditioned expert selection mechanism considering general features and domain-specific priors to dynamically choose the optimal decompression pathway. This allows task isolation and adaptive expert scaling. 
% %
% %
Last, we enhance C-MoE with the grouped dynamic filtering and hierarchical MoE-in-MoE structure. Nesting dense expert groups within sparse MoE and enabling parallel execution can reduce parameter overhead by approximately 75\% while preserving the sparse activation benefits of MoE.

Our main contributions can be summarized as follows:	
\begin{itemize}
\item To the best of our knowledge, we are the first to propose an efficient, unified  anomaly detection framework that uses a single set of parameters to handle  multi-modal  and  multi-class data.			%
\item We propose a ``general → specific'' paradigm to learn a general multi-modal encoder and feature compression module for effective cross-modal representation.
\item We design the C-MoE, which effectively reduces interference caused by domain gap, while improving the parameter efficiency of MoE.	
\item UniMMAD shares all parameters and dynamically activates domain-specific experts across diverse scenarios, surpassing state-of-the-art methods  with strong continual learning capability and high inference efficiency.
\end{itemize}

\begin{figure*}[t] 
    \centering
    \includegraphics[width=1.0\textwidth]{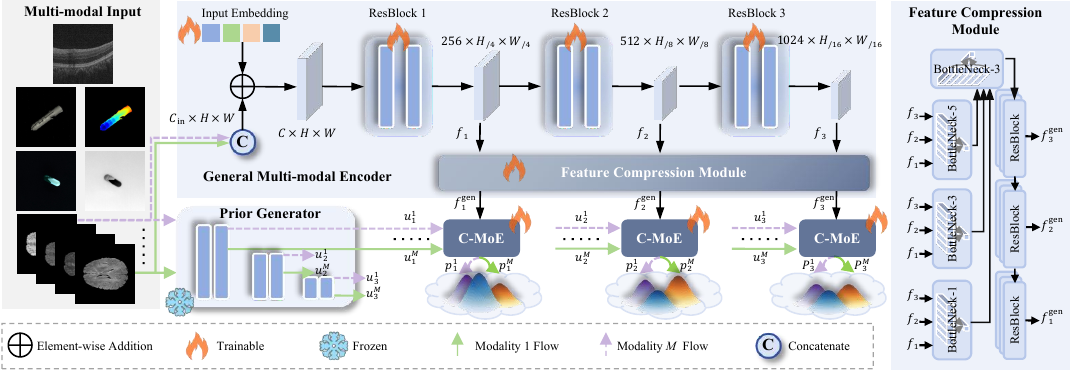}
    % \vspace{-2mm}
    \caption{
    Overview of the UniMMAD. It processes various modality combinations via a general multi-modal encoder and a Feature Compression Module (FCM). The FCM comprises a hierarchical BottleNeck-$K$ structure and residual  blocks (ResBlock), where BottleNeck-$K$ uses 
$K\times K$ convolutions to capture scale information and  $1\times 1$ convolutions to adjust dimensions. A prior generator provides domain-specific priors to guide the C-MoE  in decompressing general features into domain-specific ones.
    }
    \label{fig:frameworks}
    % \vspace{-2mm}
\end{figure*}

\section{Related Work}

\subsection{Multi-modal Anomaly Detection}
Multi-modal anomaly detection integrates complementary information from sensors such as RGB, depth, and surface normals to enhance robustness in complex environments.
M3DM~\cite{wang2023multimodal} uses patch-level contrastive learning and decision-level fusion for RGB and point clouds.
MMRD~\cite{gu2024rethinking} introduces the surface normal modality via reverse distillation.
To reduce memory usage, CFM~\cite{costanzino2024multimodal} proposes lightweight cross-modal mapping and layer pruning.
Different from these methods that train per domain and rely on parameter-free fusion, we propose a ``general $\rightarrow$ specific'' paradigm with a general encoder, enabling flexible modality combinations and deep cross-modal integration.

\subsection{Unified Vision Model}
Large-scale foundation models have driven efforts to unify diverse vision tasks with shared parameters.
In supervised settings, SegGPT~\cite{wang2023seggpt} frames segmentation as in-context coloring, and Spider~\cite{zhao2024spider} extends this to unify context-independent and medical tasks.
For anomaly detection, most methods follow a one-class-one-model scheme.
RD~\cite{deng2022anomaly} uses reverse knowledge distillation, while PatchCore~\cite{roth2022towards} compares memory bank features.
UniAD~\cite{you2022unified} proposes a unified multi-class model within a single-modality dataset.
ViTAD~\cite{joy2025vitad} and MambaAD~\cite{he2024mambaad} further improve performance via transformers and state-space models.
To address the challenges of diverse scenes, we propose a practical unified framework that supports anomaly detection across modalities, classes, and fields.	

\subsection{Mixture‑of‑Experts}
Mixture-of-Experts (MoE)~\cite{jacobs1991adaptive} leverages sparsely activated experts for scalable conditional computation.
Sparsely-Gated MoE~\cite{shazeer2017outrageously} activates a subset of experts per input, while SoftMoE~\cite{puigcerver2023sparse} improves stability via differentiable routing.
DeepSeekMoE~\cite{dai2024deepseekmoe} enhances parameter efficiency through shared experts.
In vision tasks, V-MoE~\cite{riquelme2021scaling} embeds MoE layers into ViTs.
Recent works~\cite{wang2025cnc,meng2024moead} adopt self-routing MoE structures for anomaly detection, where inputs act as both router and expert targets, leading to limited condition awareness. In this work, we propose C-MoE with a cross-condition router to enhance condition awareness by explicitly disentangling modality and class interference.

\section{Method}

The overall architecture is shown in Fig.~\ref{fig:frameworks}.
In this section, we first introduce the main idea of the “general → specific” paradigm.
Next, we describe the general multi-modal encoder  for compressing multi-modal features and the Cross Mixture-of-Experts (C-MoE)  for decompressing those features into domain-specific  ones. 
Finally, we outline the training and inference processes.

\subsection{General $ \rightarrow$ Specific }
\label{sec:share2specific}
To handle diverse modality combinations and domain discrepancies, we propose a ``general $\rightarrow$ specific'' paradigm.
Its core idea is to decompress multi-modal features into multiple uni-modal ones: $f^\text{gen} \rightarrow \{u^m\}^M_{m=1}$, where $f^\text{gen} $ is the multi-modal feature, $u^m$ is the  $m$-th uni-modal feature, and $M$ is  the number of modalities.
The model learns this decomposition on normal samples by predicting the residual between $f^\text{gen}$ and each $u^m$.
A pre-trained prior generator provides domain-specific priors to guide and supervise the decompression process, helping the model capture both general and modality-specific features.
During  inference, this decompression fails on anomalies, and the resulting deviations serve as indicators for anomaly detection.

\subsection{General Multi-modal Encoder}
\label{sec:Share_encoder}
%Todo: FCM不能称为两阶段
As shown in Fig.~\ref{fig:frameworks}, the general multi-modal encoder includes an input embedding layer, residual blocks, and a Feature Compression Module (FCM).
The input embedding layer pads all inputs to a unified channel dimension $C$, supporting arbitrary modality combinations: $\mathbb{R}^{C_\text{in}\times H\times W} \rightarrow \mathbb{R}^{C\times H\times W}$.
Three residual blocks are used to progressively extract multi-modal features, combined with inter-modal prior averages to refine the features.
%
% \noindent\textbf{Feature Compression Module.}
To prevent anomaly cues from contaminating multi-modal features, FCM employs a hierarchical bottleneck.
An inner multi-scale bottleneck    uses parallel $1\times1$, $3\times3$, and $5\times5$ convolutions to preserve normal patterns while suppressing scale-sensitive anomalies.
An outer bottleneck performs finer compression at higher semantic levels to further remove residual abnormal activations.
Finally, residual  blocks restore the compressed features to multi-scale outputs $f^\text{gen}_1$, $f^\text{gen}_2$, and $f^\text{gen}_3$, yielding purified general features.

\begin{figure*}[t] 
    \centering
    \includegraphics[width=\linewidth]{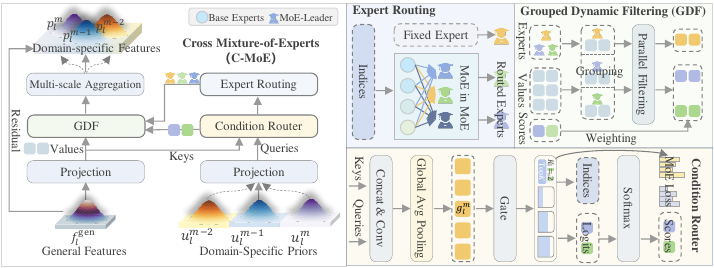}
    % \vspace{-2mm}
    \caption{
     Detailed architecture of C-MoE. It selects expert indices based on domain-specific priors using a condition router, activates the corresponding routed experts and a fixed expert, and decompresses general features via grouped dynamic filtering. Each routed expert adopts an MoE-in-MoE structure to improve parameter efficiency.
    }
    \label{fig:CrossMoE}
    % \vspace{-1mm}
\end{figure*}

\subsection{Cross Mixture-of-Experts }
\label{sec:Cross_MoE}
C-MoE decompresses general features $f^\text{gen}_l$ into domain-specific features $p^m_l$, guided by domain priors $u^m_l$ that encode multi-scale domain knowledge, with $l$ representing the index of feature maps. Instead of relying on cross attention, which may cause shortcut issues~\cite{you2022unified} and high computational overhead, C-MoE adopts a cross-condition routing strategy. It selects implicit experts based on the priors, enhancing condition awareness and reducing interference across modalities and classes. This mechanism serves as the basis for the following routing and expert design modules.

\noindent\textbf{Condition Router.}
As shown in Fig.~\ref{fig:CrossMoE}, general features are projected to keys and values, and priors to queries.
A convolutional layer followed by global average pooling yields global statistics $g^m_l$, which encapsulate domain-specific contextual characteristics and effectively suppress anomalous leakage, thereby preventing shortcut learning. 
A gating function $G$ produces top-$K$ expert indices $\mathcal{I}$ and logits, further normalized by softmax to obtain cross-condition scores.
To encourage balanced expert usage, we introduce an annealed load-balancing loss:
\begin{equation}
\mathcal{L}_\text{MoE} = \frac{1}{ML} \sum^{M}_{m=1} \sum^L_{l=1} (1 - \frac{e}{E})^2 \cdot CV(G(g^m_l)),
\end{equation}
where $e$, $E$, and $L$   represent the epoch index, total number of epochs, and the number of feature maps, respectively. The Coefficient of Variation $CV(\cdot)$ quantifies the uniformity of the gate's output distribution. This scheme encourages broad expert activation early and stabilizes routing later.

\noindent\textbf{Expert Design and Routing.}
C-MoE employs two types of experts: (1) a fixed expert to capture shared knowledge and reduce redundancy, and (2) routed experts selected via top-$K$ gating to provide task-specific capabilities.
To reduce memory cost and avoid inactive experts, we construct a MoE-in-MoE structure.
Each routed expert (MoE-Leader) is a weighted composition of shared base experts $W \in \mathbb{R}^{N_\text{exp} \times O \times I \times K_s \times K_s}$, where $N_\text{exp}$ is the number
of base experts, $O$ and $I$ represent output and input channels,
respectively, and $K_\text{s}$ is the kernel size.
Each MoE-Leader stores only weights $S \in \mathbb{R}^{N_\text{exp} \times O}$ for selecting base experts.
Given routing indices $\mathcal{I}$, final convolution kernels are computed as:
\begin{equation}
\hat{W} = \sum\nolimits_{i\in \mathcal{I}} S'[i, o] \cdot W[i, o], \quad \text{for each } o \in [1, O],
\end{equation}
where $S'$ is softmax-normalized $S$ along the first dimension.
This design enables efficient reuse of base experts, diverse MoE-Leader construction, and better parameter efficiency.

\noindent\textbf{Inference Acceleration  and Multi-scale Aggregation.} 
To accelerate C-MoE inference, we introduce two optimizations.
(1) \textit{Pre-computed kernels}. During inference, MoE-in-MoE caches the pre-weighted convolution kernels $\hat{W}$, eliminating on-the-fly kernel generation.
(2) \textit{Grouped dynamic filtering.} The values are replicated for the $K_\text{route}$   activated routed experts and one fixed expert, and reshaped from $[B=(K_\text{route}+1),C_\text{val},\cdots]$ to $[B=1,(K_\text{route}+1)\times C_\text{val},\cdots]$, where $C_\text{val}$ denotes the channel dimension of values. Setting groups = $K_\text{route}+1$ executes the  expert filters within a single group convolution, enabling parallel dynamic filtering and reducing memory traffic and latency by eliminating serial passes. 
The filtered outputs are weighted and aggregated according to the expert scores produced by the router.
To cope with objects of different spatial extents, C-MoE further employs experts with heterogeneous receptive fields  and fuses their outputs through an  aggregation convolution. %

\begin{table*}[t]
  \caption{Quantitative comparison at the image level across multi-scene datasets, with the best in \textbf{bold} and second-best \underline{underlined}.}
  % \vspace{-1.5mm}
  \label{tab:image_quanti}
  \centering
  \begin{adjustbox}{width=\linewidth}
  \begin{tabular}{llcc|cc|cc|cc|cc}
\toprule
\multirow{2}{*}{Methods} & \multirow{2}{*}{Publications} & \multicolumn{2}{c|}{MVTec-3D} & \multicolumn{2}{c|}{Eyecandies} & \multicolumn{2}{c|}{MulSen-AD} & \multicolumn{2}{c|}{BraTs}& \multicolumn{2}{c}{UniMed}  \\
\cmidrule{3-12}
 & & $\textrm{AUC}_I$ & $\textrm{MF1}_I$ & $\textrm{AUC}_I$ & $\textrm{MF1}_I$ & $\textrm{AUC}_I$ & $\textrm{MF1}_I$ & $\textrm{AUC}_I$ & $\textrm{MF1}_I$& $\textrm{AUC}_I$ & $\textrm{MF1}_I$ \\
\midrule
\multicolumn{12}{c}{\textbf{Specialized Models (One model for one dataset)}} \\
\midrule
M3DM~\cite{wang2023multimodal} & CVPR2023 & 92.129 & 94.134 & 77.387 & 76.039 & - & - & - & - & - & - \\
CFM~\cite{costanzino2024multimodal} & CVPR2024 & \uline{92.443} & \uline{93.731} & \uline{81.847} & \uline{79.668} & - & - & - & -& - & -  \\
 MulSen-TripleAD~\cite{li2024multi} & CVPR2025 & - & - & - & - & \uline{78.867} & \textbf{92.832} & - & - & - & - \\
PatchCore+MMRD & - & - & - & - & - & - & - & \uline{91.827} & \uline{90.471} & - & - \\
% \\
UniAD~\cite{you2022unified} & NeurIPS2022 & - & - & - & - &  -&  - & - & - &90.847 &	90.934 
\\
RD~\cite{deng2022anomaly} & CVPR2022 & - & - & - & - & -&  - & - & - &93.432 &	91.581 
\\
ViTAD~\cite{joy2025vitad} & CVIU2025 & - & - & - & - &  - & - & - & -& 95.524 	&93.648 
\\
MambaAD~\cite{he2024mambaad} & NeurIPS2024 & - & - & - & - & - &  - & - & -&95.977 	&93.678
 \\
SimpleNet~\cite{liu2023simplenet} & CVPR2023 & - & - & - & - &  -&  - & - & - &89.422 &	88.515 
\\
INP-Former~\cite{luo2025exploring} & CVPR2025 & - & - & - & - & -& - & - & - &\uline{96.060} &	\textbf{94.363}
\\
\midrule
\multicolumn{12}{c}{\textbf{Generalist Models (One model performs all AD tasks without fine-tuning)}} \\
\midrule
AdaCLIP~\cite{cao2024adaclip} &ECCV2024  & 74.134 & 89.929 & 73.134 & 74.809  & 59.127 & 78.126 &70.674 & 81.603 & 83.845& 91.247
 \\
MVFA~\cite{huang2024adapting} & CVPR2024 &  62.781 & 89.474 & 64.817 & 70.268& 65.781 & 79.973 & 63.749 & 79.934&88.138 &87.219
 \\
AA-CLIP~\cite{ma2025aa} & CVPR2025  & 71.827& 90.436 & 48.137 & 65.642 & 77.847 & \uline{92.261} &57.647 & 80.174&72.253& 84.193
 \\
\midrule
\multicolumn{12}{c}{ \textbf{Unified  Multi-modal and Multi-class Model}} \\
\midrule
 \rowcolor{gray!30}Ours & - & \textbf{92.527}&\textbf{94.903}&\textbf{85.566}&\textbf{84.103}&\textbf{85.460}&90.236&\textbf{95.838}&\textbf{91.139} &\textbf{96.335} &	\uline{94.182}
\\
\bottomrule
\end{tabular}

  \end{adjustbox}
  % \vspace{-1mm}
\end{table*}

\subsection{Training and Inference}
\label{sec:Training_Inference}
\noindent\textbf{Training.} We adopt a weighted sampling strategy to balance the different training tasks, where each task’s sampling probability is inversely proportional to the number of samples per class.
We introduce a decompression consistency loss $\mathcal{L}_\text{DeC}$ to enforce alignment between decompressed uni-modal features and their original counterparts, as defined below:	
\begin{equation}
\begin{aligned}
 &\mathcal{L}_\text{DeC} =  \frac{1}{M L} \sum^{M}_{m=1}\sum^L_{l=1} \sum^{H_l}_{h=1}  \sum^{W_l}_{w=1}   \frac{1}{ H_l W_l }   \text{sg}(A^\textrm{m}_l)^\gamma \odot A^\textrm{m}_l, \\
 &A^\textrm{m}_l (h,w) =  1- \frac{u^{m}_l(h,w)^T \cdot   p^m_l(h,w)}{||u^{m}_l(h,w)||\cdot||p^m_l(h,w)||}
 ,
 \end{aligned}
\end{equation}
where  $\odot$ denotes element-wise multiplication, and    $H_l$ and $W_l$ denote the height and width of the $l$-th feature map.
For the $m$-th modality at coordinate $(h,w)$ in layer $l$, the anomaly map $A^m_l$ is defined as the negative cosine similarity between the domain-specific prior $u^m_l$ and the corresponding  feature $p^m_l$ from  the C-MoE. 
The loss incorporates a focal loss~\cite{lin2017focal}–like modulation factor $\gamma$ to place greater emphasis on minority classes, where $\text{sg}(\cdot)$ denotes the stop-gradient operation.
The overall objective is end-to-end  optimized by $\mathcal{L} = \mathcal{L}_\text{DeC} +\mathcal{L}_\text{MoE}$.

\noindent\textbf{Inference.}  The model uses the discrepancy between   the decompressed uni-modal features and their original  counterparts to localize anomalies, as follows:
\begin{equation}
 \mathcal{\mathcal{S}_\textrm{AL}} =  \frac{1}{L} \sum\nolimits^L_{l=1} \Phi \left( \sqrt{\sum\nolimits^M_{m=1}  (A^\textrm{m}_l )^2}\right),
\end{equation}
where $\Phi(\cdot)$  upsamples the anomaly map to the input size and smooths it using a Gaussian  kernel with $\sigma=4$, as  in~\cite{costanzino2024multimodal}. The image-level anomaly detection score $\mathcal{S_\textrm{AD}}$ is computed as the average of the top 0.1\% of anomaly localization scores.

\begin{table*}[t]
  \caption{Quantitative comparison at the pixel level across multi-scene datasets, with the best in \textbf{bold} and second-best \underline{underlined}.}
  
  % \vspace{-1mm}
  \label{tab:pixel_quanti}
  \centering
  \begin{adjustbox}{width=0.98\linewidth}
  \begin{tabular}{llcc|cc|cc|cc|cc}
\toprule
\multirow{2}{*}{Methods} & \multirow{2}{*}{Publications} & \multicolumn{2}{c|}{MVTec-3D} & \multicolumn{2}{c|}{Eyecandies} & \multicolumn{2}{c|}{MulSen-AD} & \multicolumn{2}{c|}{BraTs}& \multicolumn{2}{c}{UniMed}  \\
\cmidrule{3-12}
 & & $\textrm{AUC}_P$ & $\textrm{MF1}_P$ & $\textrm{AUC}_P$ & $\textrm{MF1}_P$ & $\textrm{AUC}_P$ & $\textrm{MF1}_P$ & $\textrm{AUC}_P$ & $\textrm{MF1}_P$ & $\textrm{AUC}_P$ & $\textrm{MF1}_P$ \\
\midrule
\multicolumn{12}{c}{\textbf{Specialized Models  (One model for one dataset)}} \\
\midrule
M3DM~\cite{wang2023multimodal} & CVPR2023 & \uline{98.914} & \textbf{44.721} & 93.758& \uline{36.527} & - & - & - & -& - & - \\
CFM~\cite{costanzino2024multimodal} & CVPR2024 & 98.757 & 44.137 & \uline{95.838} & 31.916 & - & - & - & -& - & - \\
 MulSen-TripleAD~\cite{li2024multi} & CVPR2025 & - & - & - & - & 97.833 & \uline{33.640} & - & -& - & - \\
PatchCore+MMRD & - & - & - & - & - & - & - & 95.717 & 46.737 & - & -\\
% \\
UniAD~\cite{you2022unified} & NeurIPS2022 & - & - & - & - & - &  -& -  & - &87.020 &	39.435 
\\
RD~\cite{deng2022anomaly} & CVPR2022 & - & - & - & - &  - &  - & - & - &90.514 &	40.012 
\\
ViTAD~\cite{joy2025vitad} & CVIU2025 & - & - & - & - &  - &  -& -  & - &91.828 &	44.441 
\\
MambaAD~\cite{he2024mambaad} & NeurIPS2024 & - & - & - & - &  - &  - & - & -&91.687& 	42.401 
 \\
SimpleNet~\cite{liu2023simplenet} & CVPR2023 & - & - & - & - &  - &  - & - & -&86.858 	&39.136 \\
INP-Former~\cite{luo2025exploring} & CVPR2025 & - & - & - & - & -& - & - & - &\textbf{92.670} &	\textbf{45.477}
 \\
\midrule
\multicolumn{12}{c}{\textbf{Generalist Models (One model performs all AD tasks without fine-tuning)}} \\
\midrule
AdaCLIP~\cite{cao2024adaclip} & ECCV2024 & 96.924 & 37.106 & 95.173 &32.047 & \uline{97.835} & 29.752 & \uline{96.553} & \uline{48.439} & 90.729 & 42.852
 \\
MVFA~\cite{huang2024adapting} & CVPR2024 &93.264 & 14.012 & 82.650 & 11.105 & 90.381 & 15.817 &94.348  & 33.426&87.514 & 34.281 
 \\
AA-CLIP~\cite{ma2025aa} & CVPR2025 & 98.025 & 42.141 & 94.744& 27.205 & 97.531 & 32.501 & 94.964 & 29.991&90.943 & 40.325 
 \\
\midrule
\multicolumn{12}{c}{ \textbf{Unified  Multi-modal and Multi-class Model}} \\
\midrule
 \rowcolor{gray!30}Ours & - & 
\textbf{99.089}&\uline{44.159}&\textbf{96.913}&\textbf{39.436}&\textbf{97.919}&\textbf{34.666}&\textbf{97.468}&\textbf{51.404}&\uline{92.015}& 	\uline{44.893}
 \\
\bottomrule
\end{tabular}

  \end{adjustbox}
  % \vspace{-1mm}
\end{table*}

\section{Experiments}
\subsection{Datasets and Evaluation Metrics}
 We conduct experiments on 9 widely used AD datasets, with Hyper-Kvasir~\cite{tian2021constrained}, Retinal OCT~\cite{hu2019automated}, and Liver CT~\cite{bao2024bmad}  unified into the UniMed dataset. Detailed dataset information is summarized in the Appendix (Tab.~\ref{tab:datasets_list}). 
We follow the training settings of recent state-of-the-art methods in these tasks and merge all training samples together as our training set. 
We adopt both image-level and pixel-level metrics to evaluate all the models. For the image-level metrics, we report the Area Under the Receiver Operating Characteristic Curve ($\textrm{AUC}_I$), Average Precision ($\textrm{AP}_I$), and the maximum F1 score ($\textrm{MF1}_I$), following the methodology in the literature~\cite{defard2021padim,he2023diad}. 
For the pixel-level metric, we report the Area Under the Receiver Operating Characteristic Curve ($\textrm{AUC}_P$), Area Under the Per-Region-Overlap (AUPRO) and the maximum F1 score ($\textrm{MF1}_P$).	
$\textrm{MF1}_I$ and $\textrm{MF1}_P$ are critical metrics for evaluating anomaly detection and localization performance~\cite{jeong2023winclip}, representing the upper bounds of performance at the image and pixel levels, respectively. 

\subsection{Implementation Details}
Following common  practice~\cite{gu2024rethinking,roth2022towards} in anomaly detection, we adopt WideResNet50~\cite{zagoruyko2016wide}  as the prior generator to extract the first three layers as domain-specific priors. 
All images are resized to $256\times256$ to ensure fair comparisons.	
 We train and test    specialized  methods on their respective tasks in a conventional multi-class setting with a consistent backbone and experimental setup, while evaluating generalist methods across all datasets by predicting each modality independently and aggregating the results. 
In C-MoE, we employ 8 shared base experts and 32 MoE-Leaders, while the multi-scale experts use 
$1\times1, 3\times3$, and $5\times5$  kernels.
The  $K$ for the activated top-$K$ routed experts is set to 2. 
The decompression consistency loss uses $\gamma=2$, following focal loss~\cite{lin2017focal}. 
Additional experimental details, comparison methods and dataset descriptions can be found in the Appendix.	

\begin{figure*}[t]
  \centering
    \includegraphics[width=0.98\linewidth]{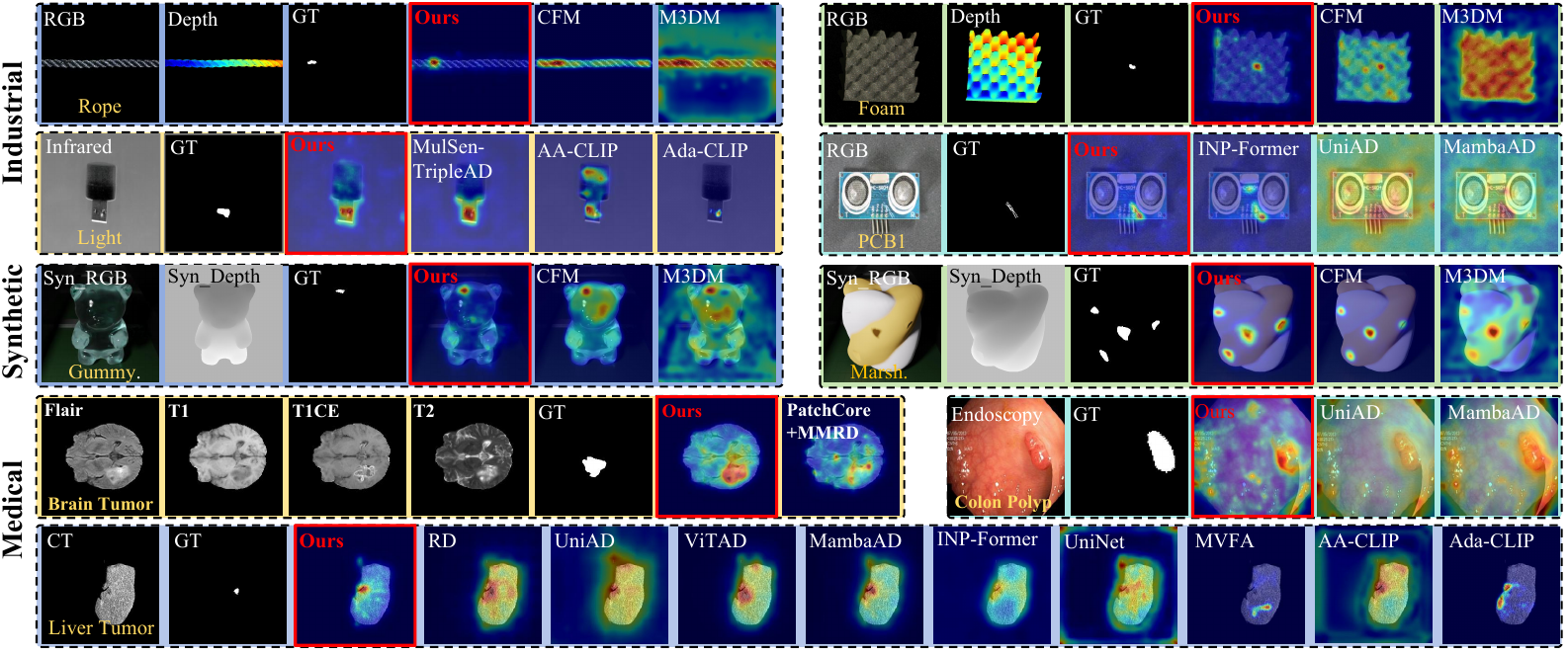}
  % \vspace{-6mm}
  \caption{Qualitative comparisons across three fields, with our method highlighted in red, modality  in white and class  in yellow.}
  \label{fig:segmentation_visualization}
  \vspace{-2mm}
\end{figure*}

\begin{table}[t]
  \caption{Quantitative comparison on the MVTec-AD and VisA datasets under the super–multi-class setting with a resolution of $256 \times 256$, evaluated using image-level metrics ($\text{AUC}_I $/$\text{AP}_I $/$ \text{MF1}_I$) and pixel-level metrics (  $\text{AUC}_P $/$\text{MF1}_P $/$ \text{AUPRO}$).
  % UniMed refers to the combination of Retinal OCT, Liver CT, and Hyper-Kvasir datasets. 
  }
  % \vspace{-2mm}
  \label{tab:mvtec_visa_multi_class}
  \centering
  \begin{adjustbox}{width=1\linewidth}
  
\begin{tabular}{l|cc|cc}
\toprule
Datasets& \multicolumn{2}{c|}{MVTec-AD}   & \multicolumn{2}{c}{VisA }   \\
\cmidrule{1-5}
Method$\downarrow{}$ & Image-level & Pixel-level & Image-level & Pixel-level \\
\midrule
RD~\cite{deng2022anomaly} & 95.8/97.8/95.0 & 95.1/51.5/90.7 & 88.4/89.9/86.7 & 96.8/38.7/87.0 \\
UniAD~\cite{you2022unified} & 95.0/97.7/94.5 & 95.8/46.8/90.0 & 90.7/92.2/86.9 & 98.3/38.1/88.8 \\
ViTAD~\cite{joy2025vitad}  & 97.7/98.9/96.5 & 97.1/57.0/90.9 & 89.1/90.4/85.4 & 98.0/39.8/84.2 \\
MambaAD~\cite{he2024mambaad}  & 94.7/97.7/94.5 & 96.3/51.5/90.5 & 90.2/91.4/87.5 & 97.7/39.4/87.9 \\
INP-Former~\cite{luo2025exploring}  & 99.2/99.5/98.6 &  \textbf{98.2/60.7/93.8} & 95.2/95.7/91.9 & 98.8/44.4/\textbf{91.5} \\
\midrule
 \rowcolor{gray!30}Ours & \textbf{99.4/99.6/98.7} & 98.1/60.2/93.0 &  \textbf{95.5/96.2/92.4} &  \textbf{98.9/47.2}/91.3 \\
\bottomrule
\end{tabular}
  \end{adjustbox}
  \vspace{-5mm}
\end{table}

\subsection{Quantitative Results}
The image-level and pixel-level results are shown in Tab.~\ref{tab:image_quanti} and Tab.~\ref{tab:pixel_quanti}, respectively.
Although trained only once on the entire dataset, UniMMAD consistently outperforms most specialized methods. It surpasses strong multi-modal baselines on MVTec-3D and  BraTS, and improves pixel-level $\text{MF1}_P$ by 7.5\% on Eyecandies.
On uni-modal tasks, generalist models transfer well to UniMed but still underperform compared with existing multi-class methods.
Furthermore, the proposed method is compared with mainstream  multi-class methods on the traditional industrial datasets MVTec-AD and VisA under the super–multi-class setting~\cite{luo2025exploring}, which constructs a unified model for  industrial scenarios. As shown in Tab.~\ref{tab:mvtec_visa_multi_class}, the proposed method outperforms current  approaches, particularly in the complex multi-instance scenes of VisA, achieving a 10.8\% improvement in $\text{MF1}_P$ than INP-Former~\cite{luo2025exploring}.
These results show that UniMMAD generalizes well across diverse AD tasks, while maintaining high accuracy without per-task customization.

\subsection{Qualitative Results}
We show some qualitative comparisons in Fig.~\ref{fig:segmentation_visualization}.
Compared to specialist methods, UniMMAD generates more precise and sharper activations on defect regions while maintaining minimal responses in normal areas. Examples include detail-rich ``Rope'', boundary-blurred ``Brain Tumor'', and low-contrast ``Light''.
Although generalist models can produce concentrated activation maps, they often highlight normal or benign regions resulting in false positives. 
Additional qualitative  discussions are provided in the Appendix.

\begin{table*}[htbp]
  \caption{Overall ablation study on the architecture, C-MoE, and training strategies.
  }
  \label{tab:ablation}
  \centering
  \begin{adjustbox}{width=1\linewidth}
  \begin{tabular}{l||ccc||ccc|ccc|ccc|ccc|ccc}
\toprule
\multirow{2}{*}{Methods} & \multicolumn{3}{c||}{\textbf{Mean}} & \multicolumn{3}{c|}{MVTec-3D} & \multicolumn{3}{c|}{Eyecandies} & \multicolumn{3}{c|}{MulSen-AD} & \multicolumn{3}{c|}{BraTs} & \multicolumn{3}{c}{UniMed}  \\
\cmidrule{2-19}
&  $\textrm{AUC}_I$ & $\textrm{AUC}_P$ & $\textrm{MF1}_P$ & $\textrm{AUC}_I$ & $\textrm{AUC}_P$ & $\textrm{MF1}_P$ & $\textrm{AUC}_I$ & $\textrm{AUC}_P$ & $\textrm{MF1}_P$ & $\textrm{AUC}_I$ & $\textrm{AUC}_P$ & $\textrm{MF1}_P$ & $\textrm{AUC}_I$ & $\textrm{AUC}_P$ & $\textrm{MF1}_P$ & $\textrm{AUC}_I$ & $\textrm{AUC}_P$ & $\textrm{MF1}_P$\\
\midrule
\multicolumn{19}{c}{\large \textbf{(a) Architecture}}
 \\
\midrule
Baseline& 
75.624&86.616&28.464&82.760&98.867&39.287&58.151&53.832&10.691&79.913&98.006&34.099&69.253&91.873&19.720&88.044&90.501&38.524
  \\
+ FCM & 77.367&86.651&28.926&84.316&98.940&39.160&57.069&52.497&9.687&80.747&98.187&33.453&73.226&92.643&21.625&91.478&90.987&40.705
  \\
+ General→Specific& 84.306&96.105&37.132&89.670&99.080&44.983&67.920&96.580&27.491&80.360&97.860&31.215&90.747&95.628&39.706&92.832&91.376&42.266
 \\
 \rowcolor{gray!30}+ C-MoE  &    91.145 & 96.681 & 42.912 & 92.527 & 99.089 & 44.159 & 85.566 & 96.913 & 39.436 & 85.460 & 97.919 & 34.666 & 95.838 & 97.468 & 51.404 & 96.335 & 92.015 & 44.893 \\
\midrule
\multicolumn{19}{c}{\large \textbf{(b) C-MoE }} \\
\midrule
 w/o  Cross-condition & 85.126 & 95.682 & 37.918 & 88.394 & 98.394 & 40.906 & 80.859 & 95.622 & 38.347 & 80.805 & 97.817 & 31.713 & 86.575 & 94.849 & 35.650 & 88.997 & 91.668 & 42.974 \\
 w/o  Multi-scale Exp. & 88.900 & 96.430 & 41.218 & 92.922 & 99.276 & 44.784 & 85.739 & 97.040 & 40.100 & 81.960 & 97.779 & 31.124 & 91.311 & 96.478 & 46.438 & 92.572 & 91.577 & 43.646 
\\
 w/o  Routed Experts & 85.417&95.967&37.822&88.275&98.431&41.369&80.000&95.646&37.040&79.750&97.645&30.014&88.010&94.594&33.583&91.051&91.875&43.355
 \\
 w/o  Fixed  Expert & 89.424 & 96.475 & 41.482 & 91.751 & 99.244 & 44.585 & 86.233 & 97.228 & 40.901 & 82.823 & 97.463 & 31.687 & 92.602 & 96.874 & 49.566 & 93.713 & 91.567 & 40.673 \\
 \rowcolor{gray!30}Ours &      91.145 & 96.681 & 42.912 & 92.527 & 99.089 & 44.159 & 85.566 & 96.913 & 39.436 & 85.460 & 97.919 & 34.666 & 95.838 & 97.468 & 51.404 & 96.335 & 92.015 & 44.893 \\

\bottomrule
\end{tabular}

  \end{adjustbox}
  % \vspace{-4mm}
\end{table*}

\subsection{Ablation Studies}
We adopt the Reverse Teacher–Student framework~\cite{deng2022anomaly} as our baseline due to its clear architecture and reliable performance, which makes it a strong foundation for extension to our unified setting. 
However, the original design is limited to single-modality input. To adapt this baseline to various modality combinations, we use a pre-trained model to extract uni-modal features and then fuse them into multi-modal representations using the parameter-free modulation  from MMRD~\cite{gu2024rethinking}.
These modifications allow the baseline to support the unified detection task and provide a fair baseline for evaluating the effectiveness of each design in UniMMAD.  Tab.~\ref{tab:ablation} presents detailed ablation results in terms of both image-level and pixel-level metrics.

\begin{figure}[htbp]
  \centering
    \includegraphics[width=\linewidth]{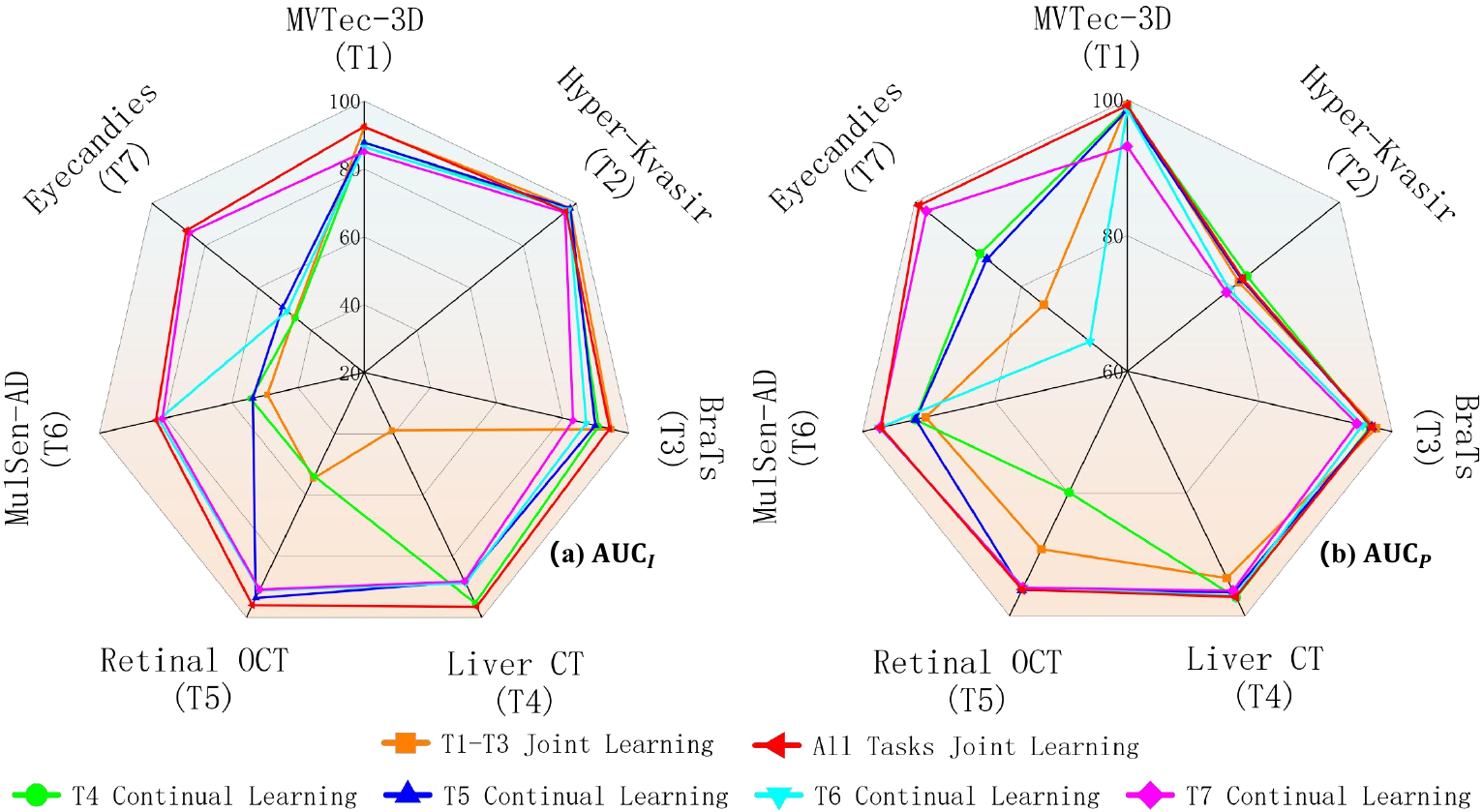}
  % \vspace{-2mm}
  \caption{ Image- / pixel-level performance of UniMMAD in continual learning on new tasks. 
 UniMMAD is initially trained jointly on tasks T1–T3, then incrementally fine-tuned on tasks T4–T7. It achieves performance comparable to training all tasks jointly.}
  \label{fig:Continual_learning_experiment}
  % \vspace{-2mm}
\end{figure}

\noindent\textbf{Architecture.}
First, we replace the OCBE module~\cite{deng2022anomaly} with FCM and achieve significant  improvements on UniMed datasets. This is because FCM compresses normal features and enlarges their separation from anomalies, making the model more sensitive to subtle defects. 
Next, we adopt a ``general $\rightarrow$ specific''  paradigm. A general encoder extracts multi-modal features, which are decompressed via cross attention. This design captures shared cues while restoring modality-specific details. It yields 8.9\% and 10.9\% gains in $\text{AUC}_I$ and $\text{AUC}_P$, respectively.
Finally, we integrate the proposed C-MoE and improve $\text{AUC}_I$ by an average of 8.1\% across the industrial, synthetic, and medical fields.
In summary, Tab.~\ref{tab:ablation} validates the effectiveness of each component without exception.

\noindent\textbf{C-MoE.} 
The core designs of C-MoE are cross-condition mechanisms  and routed experts.
Replacing the cross-condition MoE with a plain MoE leads to an average drop of 6.7\% in terms of  $\text{AUC}_I$.
Removing  routed experts implicitly disables the cross-condition mechanism, leading to a 5.7\% drop in $\text{AUC}_I$.
%
% These results demonstrate the importance of the cross-condition mechanism in selecting domain-specific expert pathways for unified anomaly detection.	
%
Besides, both multi-scale and fixed experts contribute to further performance improvements.	
We can see that multi-scale experts help capture large tissue anomalies and small tumors, improving $\text{AUC}_I$ by 4.6\% on UniMed.	

\noindent\textbf{Continual Learning.}
Fig.~\ref{fig:Continual_learning_experiment} shows the  continual learning ability of UniMMAD.
First, we jointly train tasks T1-T3 to obtain a basic general detection capability. 
Then, we progressively conduct the continual learning from T4 to T7, where we only fine-tune less than 10\% parameters from the MoE-leader, condition router, and aggregation convolution. 
To  retain routing accuracy for early tasks, 1\% of the previous data is mixed in during the training of each new task.
Finally, UniMMAD achieves comparable performance on new tasks to joint training on all tasks, while the performance degradation on previous tasks is less than 8\%.

\begin{table}[t]
  % \vspace{-2mm}
  \captionsetup{font=scriptsize} % 调整 caption 字号
  \caption{Comparison ($\textrm{AUC}_I$/$\textrm{AUC}_P$/$\textrm{MF1}_P$) with Gains ($\Delta$) between Specialized (Speci.) and Unified  Training for our UniMMAD and Baseline.}
  \label{tab:training_strategy}

  % \vspace{-3mm}
  \centering
  \begin{adjustbox}{width=1\linewidth}
    \begin{tabular}{l|c|cccccc}
\toprule
Methods & \textbf{Mean} & \textbf{MVTec-3D} & \textbf{Eyecandies} & \textbf{MulSen-AD} & \textbf{BraTs} & \textbf{UniMed} \\

 \midrule
Baseline/Speci. & 81.2/89.9/33.0 & 87.2/99.1/42.4 & 61.4/65.5/11.0 & 77.8/98.1/32.8 & 88.3/96.0/36.0 & 91.6/90.8/43.1 \\
Baseline/Unified  & 75.6/86.6/28.4 & 82.7/98.8/39.2 & 58.1/53.8/10.6 & 79.9/98.0/34.0 & 69.2/91.8/19.7 & 88.0/90.5/38.5 \\
 \rowcolor{gray!30} $\Delta$ Speci.-Unified & \reddown5.6/\reddown3.3/\reddown4.6 & \reddown4.5/\reddown0.3/\reddown3.2 & \reddown3.3/\reddown11.7/\reddown0.4 & \greenup2.1/\reddown0.1/\greenup1.2 & \reddown19.1/\reddown4.2/\reddown16.3 & \reddown3.6/\reddown0.3/\reddown4.6 \\
\midrule
Ours/Speci. & 91.2/96.7/43.2 & 92.9/99.1/45.6 & 86.7/97.1/40.8 & 83.4/97.5/34.3 & 95.0/97.2/49.9 & 97.7/92.6/45.5 \\
 Ours/Unified & 91.1/96.6/42.9 & 92.5/99.0/44.1 & 85.5/96.9/39.4 & 85.4/97.9/34.6 & 95.8/97.4/51.4 & 96.3/92.0/44.8 \\
 \rowcolor{gray!30} $\Delta$ Speci.-Unified& \reddown0.1/\reddown0.1/\reddown0.3 & \reddown0.4/\reddown0.1/\reddown1.5 & \reddown1.2/\reddown0.2/\reddown1.4 & \greenup2.0/\greenup0.4/\greenup0.3 & \greenup0.8/\greenup0.2/\greenup1.5 & \reddown1.4/\reddown0.6/\reddown0.7 \\
\bottomrule
\end{tabular}

  \end{adjustbox} 
  % \vspace{-5mm}
\end{table}

\begin{figure}[t]
  \centering
    \includegraphics[width=\linewidth] {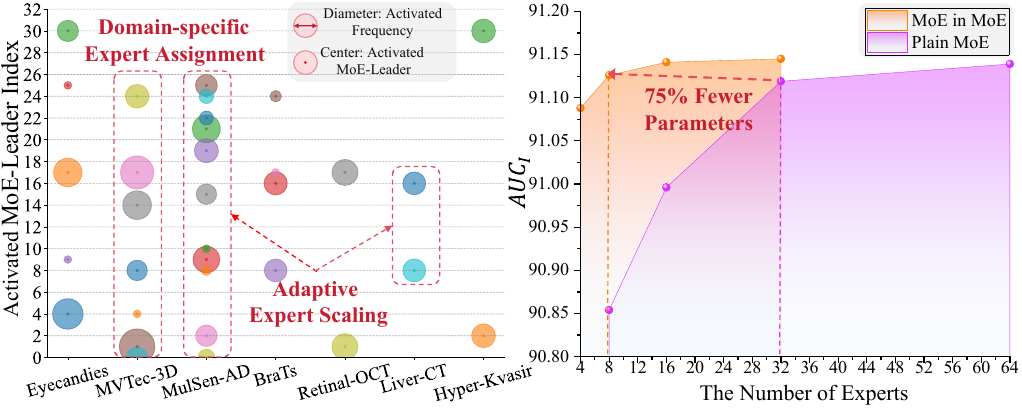}
  % \vspace{-1mm}
  \caption{ (a) Left: Activation frequency of each MoE-leader across different datasets, illustrating domain-specific expert selection. (b) Right: Comparison of  parameter efficiency   between MoE-in-MoE and plain MoE on performance for different numbers of experts. }
  \label{fig:MInMoE_And_Activation}
  % \vspace{-1mm}
\end{figure}

\noindent\textbf{Task isolation within MoE.}
Fig.~\ref{fig:MInMoE_And_Activation}(a) shows that C‑MoE assigns distinct experts to different domains, mitigating interference across modalities and classes.  
C‑MoE also dynamically adjusts the number of selected experts to match dataset complexity.  
For example, MulSen-AD exhibits high intra-class variability, while MVTec-3D introduces even more complex modality variations, thereby requiring more experts.
Compared to fixed reconstruction pathways, C‑MoE offers greater flexibility and scalability.

\noindent\textbf{MoE in  MoE.} 
To evaluate parameter efficiency, we compared MoE‑in‑MoE with plain MoE under different expert configurations.
The number of MoE‑in‑MoE experts is fixed at 32.
The number of base experts serving as the primary source of parameters is variable.
Fig.~\ref{fig:MInMoE_And_Activation} (b)  shows that MoE-in-MoE achieves performance comparable to the plain MoE while using 75\% fewer parameters.
With the increasing number of parameters, the performance gains of MoE-in-MoE gradually saturate.

\noindent\textbf{Unified vs. Specialized.} 
 Unlike baselines suffering from parameter interference, UniMMAD maintains exceptional stability (Tab.~\ref{tab:training_strategy}) by effectively disentangling domain conflicts via adaptive routing. Additionally, unified training yields the gains on data-scarce infrared modalities (MulSen-AD) and   complex 4-modality BraTs. 

\begin{table}[t]
  \caption{	 Efficiency comparison on NVIDIA 4090.   ``–'' represents memory bank–based methods without parameters(Param.), activated parameters(ActPar.) and FLOPs.}
  % \vspace{-1mm}
  \label{tab:speed_memory}
  \centering
  \begin{adjustbox}{width=1\linewidth}
  \begin{tabular}{l|rrrrr}
\toprule
\multirow{2}{*}{\textbf{Methods}} & \textbf{FPS}  & \textbf{Memory}& \textbf{Param.}& \textbf{ActPar.}& \textbf{FLOPs}\\
   & \textbf{(fps)$\uparrow$}  & \textbf{(M)$\downarrow$ } & \textbf{(MB)$\downarrow$} & \textbf{(MB)$\downarrow$} & \textbf{(GFlops)$\downarrow$}  \\

\midrule
M3DM~\cite{wang2023multimodal}        & 0.39         & 73774.53 & --& -- & --\\
 MulSen.~\cite{li2024multi}   & 1.19         & 10323.05 & -- & --& --\\
CFM~\cite{costanzino2024multimodal}         & 13.18        & \textbf{3757.12} & \textbf{113.17} &\textbf{113.17}&431.14\\
 \rowcolor{gray!30}Ours        & \textbf{59.09} & 4970.17 & 233.66 &192.38&\textbf{110.91}  \\
\bottomrule
\end{tabular}

  \end{adjustbox} 
  % \vspace{-1mm}
\end{table}

\noindent\textbf{Efficiency.}  
As shown in Tab.~\ref{tab:speed_memory}, the proposed model integrates multi-field, multi-modal, and multi-class data with a parameter count comparable to specialized multi-modal methods. Memory bank-based methods such as M3DM and MulSen-TripleAD (MulSen.) show slower inference and high memory overhead, while CFM also suffers from inefficiency due to the heavy Transformer architecture and point cloud processing. In contrast, our fully convolutional MoE-in-MoE design reuses base experts, yielding faster inference, making it well suited for practical anomaly detection.

\section{Conclusion}
We introduce UniMMAD, a unified model designed for   multi-modal and multi-class anomaly detection.
It incorporates a ``general $ \rightarrow$ specific'' paradigm, a general multi-modal encoder, and a parameter-efficient C-MoE.
 Extensive experiments on nine challenging benchmarks demonstrate that UniMMAD establishes a new state-of-the-art, outperforming specialized methods using only a single set of parameters. Furthermore, UniMMAD provides a strong baseline for unified multi-scene anomaly detection, while its high inference efficiency and continual learning capability underscore its scalability and practicality for real-world deployment.

% \section*{Acknowledgments}
% This research was supported by the National Natural Science Foundation of China under Grant 62431004, Dalian Science and Technology Innovation Foundation under Grant 2023JJ12GX015, NTU AI4X Postdoctoral Grant funded by the National Research Foundation, Singapore.

\clearpage
\appendix
\bibliography{main}

\begin{thebibliography}{43}
\providecommand{\natexlab}[1]{#1}
\providecommand{\url}[1]{\texttt{#1}}
\expandafter\ifx\csname urlstyle\endcsname\relax
  \providecommand{\doi}[1]{doi: #1}\else
  \providecommand{\doi}{doi: \begingroup \urlstyle{rm}\Url}\fi

\bibitem[Bergmann et~al.(2019)Bergmann, Fauser, Sattlegger, and
  Steger]{bergmann2019mvtec}
Paul Bergmann, Michael Fauser, David Sattlegger, and Carsten Steger.
\newblock Mvtec ad--a comprehensive real-world dataset for unsupervised anomaly
  detection.
\newblock In \emph{Proceedings of the IEEE/CVF conference on computer vision
  and pattern recognition}, pages 9592--9600, 2019.

\bibitem[Li et~al.(2018)Li, Wang, Han, Xue, Wei, Li, and
  Fei-Fei]{li2018thoracic}
Zhe Li, Chong Wang, Mei Han, Yuan Xue, Wei Wei, Li-Jia Li, and Li~Fei-Fei.
\newblock Thoracic disease identification and localization with limited
  supervision.
\newblock In \emph{Proceedings of the IEEE conference on computer vision and
  pattern recognition}, pages 8290--8299, 2018.

\bibitem[Costanzino et~al.(2024)Costanzino, Ramirez, Lisanti, and
  Di~Stefano]{costanzino2024multimodal}
Alex Costanzino, Pierluigi~Zama Ramirez, Giuseppe Lisanti, and Luigi
  Di~Stefano.
\newblock Multimodal industrial anomaly detection by crossmodal feature
  mapping.
\newblock In \emph{Proceedings of the IEEE/CVF Conference on Computer Vision
  and Pattern Recognition}, pages 17234--17243, 2024.

\bibitem[Li et~al.(2025)Li, Zheng, Xu, Gan, Lu, Li, Ni, Tian, Huang, Gao,
  et~al.]{li2024multi}
Wenqiao Li, Bozhong Zheng, Xiaohao Xu, Jinye Gan, Fading Lu, Xiang Li, Na~Ni,
  Zheng Tian, Xiaonan Huang, Shenghua Gao, et~al.
\newblock Multi-sensor object anomaly detection: Unifying appearance, geometry,
  and internal properties.
\newblock In \emph{Proceedings of the Computer Vision and Pattern Recognition
  Conference}, pages 9984--9993, 2025.

\bibitem[Wang et~al.(2023{\natexlab{a}})Wang, Peng, Zhang, Yi, Wang, and
  Wang]{wang2023multimodal}
Yue Wang, Jinlong Peng, Jiangning Zhang, Ran Yi, Yabiao Wang, and Chengjie
  Wang.
\newblock Multimodal industrial anomaly detection via hybrid fusion.
\newblock In \emph{Proceedings of the IEEE/CVF Conference on Computer Vision
  and Pattern Recognition}, pages 8032--8041, 2023{\natexlab{a}}.

\bibitem[You et~al.(2022)You, Cui, Shen, Yang, Lu, Zheng, and
  Le]{you2022unified}
Zhiyuan You, Lei Cui, Yujun Shen, Kai Yang, Xin Lu, Yu~Zheng, and Xinyi Le.
\newblock A unified model for multi-class anomaly detection.
\newblock \emph{Advances in Neural Information Processing Systems},
  35:\penalty0 4571--4584, 2022.

\bibitem[He et~al.(2024)He, Bai, Zhang, He, Chen, Gan, Wang, Li, Tian, and
  Xie]{he2024mambaad}
Haoyang He, Yuhu Bai, Jiangning Zhang, Qingdong He, Hongxu Chen, Zhenye Gan,
  Chengjie Wang, Xiangtai Li, Guanzhong Tian, and Lei Xie.
\newblock Mambaad: Exploring state space models for multi-class unsupervised
  anomaly detection.
\newblock \emph{arXiv preprint arXiv:2404.06564}, 2024.

\bibitem[Weninger et~al.(2018)Weninger, Rippel, Koppers, and
  Merhof]{weninger2018segmentation}
Leon Weninger, Oliver Rippel, Simon Koppers, and Dorit Merhof.
\newblock Segmentation of brain tumors and patient survival prediction: Methods
  for the brats 2018 challenge.
\newblock In \emph{International MICCAI brainlesion workshop}, pages 3--12.
  Springer, 2018.

\bibitem[Luo et~al.(2025)Luo, Cao, Yao, Zhang, Lou, Cheng, Shen, and
  Yu]{luo2025exploring}
Wei Luo, Yunkang Cao, Haiming Yao, Xiaotian Zhang, Jianan Lou, Yuqi Cheng,
  Weiming Shen, and Wenyong Yu.
\newblock Exploring intrinsic normal prototypes within a single image for
  universal anomaly detection.
\newblock In \emph{Proceedings of the Computer Vision and Pattern Recognition
  Conference}, pages 9974--9983, 2025.

\bibitem[Zhao et~al.(2024)Zhao, Pang, Ji, Sheng, Zuo, Zhang, and
  Lu]{zhao2024spider}
Xiaoqi Zhao, Youwei Pang, Wei Ji, Baicheng Sheng, Jiaming Zuo, Lihe Zhang, and
  Huchuan Lu.
\newblock Spider: A unified framework for context-dependent concept
  segmentation.
\newblock \emph{arXiv preprint arXiv:2405.01002}, 2024.

\bibitem[Wang et~al.(2023{\natexlab{b}})Wang, Zhang, Cao, Wang, Shen, and
  Huang]{wang2023seggpt}
Xinlong Wang, Xiaosong Zhang, Yue Cao, Wen Wang, Chunhua Shen, and Tiejun
  Huang.
\newblock Seggpt: Segmenting everything in context.
\newblock \emph{arXiv preprint arXiv:2304.03284}, 2023{\natexlab{b}}.

\bibitem[Potlapalli et~al.(2023)Potlapalli, Zamir, Khan, and
  Shahbaz~Khan]{potlapalli2023promptir}
Vaishnav Potlapalli, Syed~Waqas Zamir, Salman~H Khan, and Fahad Shahbaz~Khan.
\newblock Promptir: Prompting for all-in-one image restoration.
\newblock \emph{Advances in Neural Information Processing Systems},
  36:\penalty0 71275--71293, 2023.

\bibitem[Liu et~al.(2024)Liu, Feng, Xue, Wang, Wu, Lu, Zhao, Deng, Zhang, Ruan,
  et~al.]{liu2024deepseek}
Aixin Liu, Bei Feng, Bing Xue, Bingxuan Wang, Bochao Wu, Chengda Lu, Chenggang
  Zhao, Chengqi Deng, Chenyu Zhang, Chong Ruan, et~al.
\newblock Deepseek-v3 technical report.
\newblock \emph{arXiv preprint arXiv:2412.19437}, 2024.

\bibitem[Gu et~al.(2024)Gu, Zhang, Liu, Chen, Peng, Gan, Jiang, Shu, Wang, and
  Ma]{gu2024rethinking}
Zhihao Gu, Jiangning Zhang, Liang Liu, Xu~Chen, Jinlong Peng, Zhenye Gan,
  Guannan Jiang, Annan Shu, Yabiao Wang, and Lizhuang Ma.
\newblock Rethinking reverse distillation for multi-modal anomaly detection.
\newblock In \emph{Proceedings of the AAAI Conference on Artificial
  Intelligence}, volume~38, pages 8445--8453, 2024.

\bibitem[Deng and Li(2022)]{deng2022anomaly}
Hanqiu Deng and Xingyu Li.
\newblock Anomaly detection via reverse distillation from one-class embedding.
\newblock In \emph{Proceedings of the IEEE/CVF Conference on Computer Vision
  and Pattern Recognition}, pages 9737--9746, 2022.

\bibitem[Roth et~al.(2022)Roth, Pemula, Zepeda, Sch{\"o}lkopf, Brox, and
  Gehler]{roth2022towards}
Karsten Roth, Latha Pemula, Joaquin Zepeda, Bernhard Sch{\"o}lkopf, Thomas
  Brox, and Peter Gehler.
\newblock Towards total recall in industrial anomaly detection.
\newblock In \emph{Proceedings of the IEEE/CVF Conference on Computer Vision
  and Pattern Recognition}, pages 14318--14328, 2022.

\bibitem[Joy et~al.(2025)Joy, Nasrin, Siddiqua, and Farid]{joy2025vitad}
Md~Ashif~Mahmud Joy, Shamima Nasrin, Ayesha Siddiqua, and Dewan~Md Farid.
\newblock Vitad: Leveraging modified vision transformer for alzheimer’s
  disease multi-stage classification from brain mri scans.
\newblock \emph{Brain Research}, 1847:\penalty0 149302, 2025.

\bibitem[Jacobs et~al.(1991)Jacobs, Jordan, Nowlan, and
  Hinton]{jacobs1991adaptive}
Robert~A Jacobs, Michael~I Jordan, Steven~J Nowlan, and Geoffrey~E Hinton.
\newblock Adaptive mixtures of local experts.
\newblock \emph{Neural computation}, 3\penalty0 (1):\penalty0 79--87, 1991.

\bibitem[Shazeer et~al.(2016)Shazeer, Mirhoseini, Maziarz, Davis, Le, Hinton,
  and Dean]{shazeer2017outrageously}
Noam Shazeer, Azalia Mirhoseini, Krzysztof Maziarz, Andy Davis, Quoc~V Le,
  Geoffrey Hinton, and Jeff Dean.
\newblock Outrageously large neural networks: The sparsely-gated
  mixture-of-experts layer.
\newblock In \emph{International Conference on Learning Representations
  (2016)}, 2016.
\newblock URL \url{https://openreview.net/forum?id=B1ckMDqlg}.

\bibitem[Puigcerver et~al.(2023)Puigcerver, Riquelme, Mustafa, and
  Houlsby]{puigcerver2023sparse}
Joan Puigcerver, Carlos Riquelme, Basil Mustafa, and Neil Houlsby.
\newblock From sparse to soft mixtures of experts.
\newblock \emph{arXiv preprint arXiv:2308.00951}, 2023.

\bibitem[Dai et~al.(2024)Dai, Deng, Zhao, Xu, Gao, Chen, Li, Zeng, Yu, Wu,
  et~al.]{dai2024deepseekmoe}
Damai Dai, Chengqi Deng, Chenggang Zhao, RX~Xu, Huazuo Gao, Deli Chen, Jiashi
  Li, Wangding Zeng, Xingkai Yu, Yu~Wu, et~al.
\newblock Deepseekmoe: Towards ultimate expert specialization in
  mixture-of-experts language models.
\newblock \emph{arXiv preprint arXiv:2401.06066}, 2024.

\bibitem[Riquelme et~al.(2021)Riquelme, Puigcerver, Mustafa, Neumann, Jenatton,
  Susano~Pinto, Keysers, and Houlsby]{riquelme2021scaling}
Carlos Riquelme, Joan Puigcerver, Basil Mustafa, Maxim Neumann, Rodolphe
  Jenatton, Andr{\'e} Susano~Pinto, Daniel Keysers, and Neil Houlsby.
\newblock Scaling vision with sparse mixture of experts.
\newblock \emph{Advances in Neural Information Processing Systems},
  34:\penalty0 8583--8595, 2021.

\bibitem[Wang et~al.(2025)Wang, Wang, Bai, Lim, and Xiao]{wang2025cnc}
Xiaolei Wang, Xiaoyang Wang, Huihui Bai, Eng~Gee Lim, and Jimin Xiao.
\newblock Cnc: Cross-modal normality constraint for unsupervised multi-class
  anomaly detection.
\newblock In \emph{Proceedings of the AAAI Conference on Artificial
  Intelligence}, volume~39, pages 7943--7951, 2025.

\bibitem[Meng et~al.(2024)Meng, Meng, Zhou, Li, Hou, and He]{meng2024moead}
Shiyuan Meng, Wenchao Meng, Qihang Zhou, Shizhong Li, Weiye Hou, and Shibo He.
\newblock Moead: A parameter-efficient model for multi-class anomaly detection.
\newblock In \emph{European Conference on Computer Vision}, pages 345--361.
  Springer, 2024.

\bibitem[Liu et~al.(2023)Liu, Zhou, Xu, and Wang]{liu2023simplenet}
Zhikang Liu, Yiming Zhou, Yuansheng Xu, and Zilei Wang.
\newblock Simplenet: A simple network for image anomaly detection and
  localization.
\newblock In \emph{Proceedings of the IEEE/CVF Conference on Computer Vision
  and Pattern Recognition}, pages 20402--20411, 2023.

\bibitem[Cao et~al.(2024)Cao, Zhang, Frittoli, Cheng, Shen, and
  Boracchi]{cao2024adaclip}
Yunkang Cao, Jiangning Zhang, Luca Frittoli, Yuqi Cheng, Weiming Shen, and
  Giacomo Boracchi.
\newblock Adaclip: Adapting clip with hybrid learnable prompts for zero-shot
  anomaly detection.
\newblock In \emph{European Conference on Computer Vision}, pages 55--72.
  Springer, 2024.

\bibitem[Huang et~al.(2024)Huang, Jiang, Feng, Zhang, Wang, and
  Wang]{huang2024adapting}
Chaoqin Huang, Aofan Jiang, Jinghao Feng, Ya~Zhang, Xinchao Wang, and Yanfeng
  Wang.
\newblock Adapting visual-language models for generalizable anomaly detection
  in medical images.
\newblock In \emph{Proceedings of the IEEE/CVF Conference on Computer Vision
  and Pattern Recognition}, pages 11375--11385, 2024.

\bibitem[Ma et~al.(2025)Ma, Zhang, Yao, Tang, Wu, Li, Yan, Jiang, and
  Zhou]{ma2025aa}
Wenxin Ma, Xu~Zhang, Qingsong Yao, Fenghe Tang, Chenxu Wu, Yingtai Li, Rui Yan,
  Zihang Jiang, and S~Kevin Zhou.
\newblock Aa-clip: Enhancing zero-shot anomaly detection via anomaly-aware
  clip.
\newblock In \emph{Proceedings of the Computer Vision and Pattern Recognition
  Conference}, pages 4744--4754, 2025.

\bibitem[Lin et~al.(2017)Lin, Goyal, Girshick, He, and
  Doll{\'a}r]{lin2017focal}
Tsung-Yi Lin, Priya Goyal, Ross Girshick, Kaiming He, and Piotr Doll{\'a}r.
\newblock Focal loss for dense object detection.
\newblock In \emph{Proceedings of the IEEE international conference on computer
  vision}, pages 2980--2988, 2017.

\bibitem[Tian et~al.(2021)Tian, Pang, Liu, Chen, Shin, Verjans, Singh, and
  Carneiro]{tian2021constrained}
Yu~Tian, Guansong Pang, Fengbei Liu, Yuanhong Chen, Seon~Ho Shin, Johan~W
  Verjans, Rajvinder Singh, and Gustavo Carneiro.
\newblock Constrained contrastive distribution learning for unsupervised
  anomaly detection and localisation in medical images.
\newblock In \emph{Medical Image Computing and Computer Assisted
  Intervention--MICCAI 2021: 24th International Conference, Strasbourg, France,
  September 27--October 1, 2021, Proceedings, Part V 24}, pages 128--140.
  Springer, 2021.

\bibitem[Hu et~al.(2019)Hu, Chen, and Yi]{hu2019automated}
Junjie Hu, Yuanyuan Chen, and Zhang Yi.
\newblock Automated segmentation of macular edema in oct using deep neural
  networks.
\newblock \emph{Medical image analysis}, 55:\penalty0 216--227, 2019.

\bibitem[Bao et~al.(2024)Bao, Sun, Deng, He, Zhang, and Li]{bao2024bmad}
Jinan Bao, Hanshi Sun, Hanqiu Deng, Yinsheng He, Zhaoxiang Zhang, and Xingyu
  Li.
\newblock Bmad: Benchmarks for medical anomaly detection.
\newblock In \emph{Proceedings of the IEEE/CVF Conference on Computer Vision
  and Pattern Recognition}, pages 4042--4053, 2024.

\bibitem[Defard et~al.(2021)Defard, Setkov, Loesch, and
  Audigier]{defard2021padim}
Thomas Defard, Aleksandr Setkov, Angelique Loesch, and Romaric Audigier.
\newblock Padim: a patch distribution modeling framework for anomaly detection
  and localization.
\newblock In \emph{International Conference on Pattern Recognition}, pages
  475--489. Springer, 2021.

\bibitem[He et~al.(2023)He, Zhang, Chen, Chen, Li, Chen, Wang, Wang, and
  Xie]{he2023diad}
Haoyang He, Jiangning Zhang, Hongxu Chen, Xuhai Chen, Zhishan Li, Xu~Chen,
  Yabiao Wang, Chengjie Wang, and Lei Xie.
\newblock Diad: A diffusion-based framework for multi-class anomaly detection.
\newblock \emph{arXiv preprint arXiv:2312.06607}, 2023.

\bibitem[Jeong et~al.(2023)Jeong, Zou, Kim, Zhang, Ravichandran, and
  Dabeer]{jeong2023winclip}
Jongheon Jeong, Yang Zou, Taewan Kim, Dongqing Zhang, Avinash Ravichandran, and
  Onkar Dabeer.
\newblock Winclip: Zero-/few-shot anomaly classification and segmentation.
\newblock In \emph{Proceedings of the IEEE/CVF Conference on Computer Vision
  and Pattern Recognition}, pages 19606--19616, 2023.

\bibitem[Zagoruyko and Komodakis(2016)]{zagoruyko2016wide}
Sergey Zagoruyko and Nikos Komodakis.
\newblock Wide residual networks.
\newblock In \emph{Procedings of the British Machine Vision Conference 2016}.
  British Machine Vision Association, 2016.

\bibitem[bra(2020)]{brats2020}
Brats 2020: Brain tumor segmentation challenge.
\newblock \url{https://www.med.upenn.edu/cbica/brats2020/data.html}, 2020.

\bibitem[Fan et~al.(2025)Fan, Huang, Di, Su, Song, Pagnucco, and
  Song]{fan2025salvaging}
Lei Fan, Junjie Huang, Donglin Di, Anyang Su, Tianyou Song, Maurice Pagnucco,
  and Yang Song.
\newblock Salvaging the overlooked: Leveraging class-aware contrastive learning
  for multi-class anomaly detection.
\newblock In \emph{Proceedings of the IEEE/CVF International Conference on
  Computer Vision}, pages 21419--21428, 2025.

\bibitem[Lepikhin et~al.(2020)Lepikhin, Lee, Xu, Chen, Firat, Huang, Krikun,
  Shazeer, and Chen]{lepikhin2020gshard}
Dmitry Lepikhin, HyoukJoong Lee, Yuanzhong Xu, Dehao Chen, Orhan Firat, Yanping
  Huang, Maxim Krikun, Noam Shazeer, and Zhifeng Chen.
\newblock Gshard: Scaling giant models with conditional computation and
  automatic sharding.
\newblock \emph{arXiv preprint arXiv:2006.16668}, 2020.

\bibitem[Fedus et~al.(2022)Fedus, Zoph, and Shazeer]{fedus2022switch}
William Fedus, Barret Zoph, and Noam Shazeer.
\newblock Switch transformers: Scaling to trillion parameter models with simple
  and efficient sparsity.
\newblock \emph{Journal of Machine Learning Research}, 23\penalty0
  (120):\penalty0 1--39, 2022.

\bibitem[Bergmann et~al.(2021)Bergmann, Jin, Sattlegger, and
  Steger]{bergmann2021mvtec}
Paul Bergmann, Xin Jin, David Sattlegger, and Carsten Steger.
\newblock The mvtec 3d-ad dataset for unsupervised 3d anomaly detection and
  localization.
\newblock \emph{arXiv preprint arXiv:2112.09045}, 2021.

\bibitem[Bonfiglioli et~al.(2022)Bonfiglioli, Toschi, Silvestri, Fioraio, and
  De~Gregorio]{bonfiglioli2022eyecandies}
Luca Bonfiglioli, Marco Toschi, Davide Silvestri, Nicola Fioraio, and Daniele
  De~Gregorio.
\newblock The eyecandies dataset for unsupervised multimodal anomaly detection
  and localization.
\newblock In \emph{Proceedings of the Asian Conference on Computer Vision},
  pages 3586--3602, 2022.

\bibitem[Zou et~al.(2022)Zou, Jeong, Pemula, Zhang, and Dabeer]{zou2022spot}
Yang Zou, Jongheon Jeong, Latha Pemula, Dongqing Zhang, and Onkar Dabeer.
\newblock Spot-the-difference self-supervised pre-training for anomaly
  detection and segmentation.
\newblock In \emph{European Conference on Computer Vision}, pages 392--408.
  Springer, 2022.

\end{thebibliography}

\twocolumn[%
\begin{center}
    {\bf \Huge Appendix \\[0.3em]}
\end{center}
]

\renewcommand{\labelitemi}{-}  % 将项目前的符号改为横杠
% 确保显示章节编号为字母 A, B, C, D...
\makeatletter
\renewcommand{\thesection}{\Alph{section}}  % 设置章节编号为字母
\makeatother

% 让章节编号从A开始
\setcounter{section}{0}

\section{Overview} 
The appendix includes the following sections to complement the main manuscript:
\begin{itemize}
    \item Sec.~\ref{sec:details}: Additional  Experimental Details 
    \item Sec.~\ref{sec:dataset_datails}: Dataset Details
    \item Sec.~\ref{sec:comparsion_methods_baselines}: Methods and Baselines for Comparison
    \item Sec.~\ref{sec:More Quantitative Comparison}: More Quantitative Comparison
    \item Sec.~\ref{sec:Ablation_exp_num}: Ablation on  Number of Activated Experts
    % \item Comparison of Inference Efficiency
    \item Sec.~\ref{sec:resolution_backbone}: Influence of Resolution and Backbone
    \item Sec.~\ref{sec:Per-Class Results}: Per-Class Results
    \item Sec.~\ref{sec:More Qualitative Comparison}: More Qualitative Comparison
\end{itemize}

\section{Additional  Experimental Details} 
\label{sec:details}
\noindent\textbf{Experimental Details}. 
All experiments were conducted over 300 epochs with a batch size of 10 on one NVIDIA RTX 4090 GPU.	
The number of input channels $C$ to the general multi-modal encoder is set to 4.
The Adam optimizer was used to optimize network parameters, with the model’s learning rate set to $1\times 10^{-3}$.

\begin{figure}[htbp]
  \centering
    \includegraphics[width=1.0\linewidth]{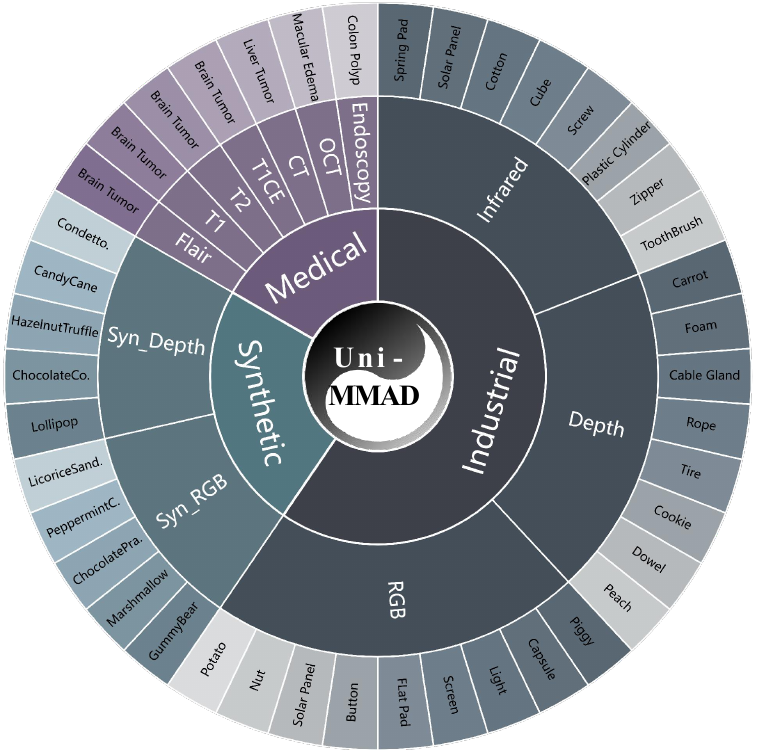}
  \caption{ Dataset details for the multi-scene, multi-modal, multi-class benchmark. From the inner ring to the outer ring, the rings represent fields, modalities, and classes, respectively.}
  \label{fig:dataset_circle_details}
\end{figure}

\section{Dataset Details} 
\label{sec:dataset_datails}
Fig.~\ref{fig:dataset_circle_details} provides an overview  of involved datasets, and  the number of images in each dataset is listed in Tab.\ref{tab:datasets_list}.
We split the BraTS 2020 dataset~\cite{brats2020} into training and test sets, including both normal and abnormal brain images. The enhancing tumor, peritumoral edema, and necrotic/non-enhancing tumor core are collectively treated as the abnormal region.
For the Retinal OCT and Liver CT datasets, we followed the BMAD protocol~\cite{bao2024bmad}   and retained only the images with clearer tissue structures.
Hyper-Kvasir, Retinal OCT, and Liver CT were grouped into a unified three-class medical dataset (UniMed).
In MulSen-AD~\cite{li2024multi}, only the infrared images were retained because the multi-modal images lacked proper calibration. 
All other datasets remain unchanged.

\section{Methods and Baselines for Comparison} 
\label{sec:comparsion_methods_baselines}
As shown in Tab.~1, We evaluate UniMMAD against SOTA specialized and generalist models across diverse domains  under the same resolution, post-processing, and evaluation metrics.
Specialized methods include M3DM, CFM,  MulSen-TripleAD, and P+MMRD (PatchCore~\cite{roth2022towards} + parameter-free fusion~\cite{gu2024rethinking}), with WideResNet50 as the RGB encoder for memory-based models.	
Uni-modal baselines include RD, SimpleNet, as well as several multi-class methods such as UniAD,  ViTAD, and MambaAD. 
All specialized models are tailored to specific modality combinations and evaluated on their respective datasets under a multi-class setting.	
We also evaluate generalist models AdaCLIP, MVFA, and AA-CLIP, which predict each modality independently and aggregate the results.
Among the specialized models, we compared four representative multi-modal approaches—M3DM, CFM, MulSen-AD, and PatchCore+MMRD (  PatchCore~\cite{roth2022towards} + the parameter-free fusion strategy from MMRD~\cite{gu2024rethinking}). For fairness, the RGB encoders of memory bank–based methods are kept consistent with our model, and both are implemented using WideResNet50.
For uni-modal methods, we consider several  widely adopted baselines,   including RD, SimpleNet, PatchCore, UniAD, ViTAD,  MambaAD, and INP-Former.
These specialized methods are tailored to specific modality combinations and are evaluated independently on their corresponding datasets at a resolution of $256 \times 256$, whereas ViT-like architectures use a resolution of $252 \times 252$.
We also evaluate recent generalist models, including AdaCLIP, MVFA, and AA-CLIP, which predict results for each modality separately and aggregate the outputs.

\begin{table*}[htbp]
  \caption{More comprehensive  comparison on the MVTec-AD and VisA datasets under the super–multi-class setting with a resolution of $256 \times 256$, evaluated using image-level metrics ($\text{AUC}_I $/$\text{AP}_I $/$ \text{MF1}_I$) and pixel-level metrics (  $\text{AUC}_P $/$\text{MF1}_P $/$ \text{AUPRO}$).
  }
  \vspace{-2mm}
  \label{tab:appendix_mvtec_visa}
  \centering
  \begin{adjustbox}{width=0.85\linewidth}
  
\begin{tabular}{ll|cc|cc}
\toprule
Datasets&Publication& \multicolumn{2}{c|}{MVTec-AD}   & \multicolumn{2}{c}{VisA }   \\
\cmidrule{1-6}
Method$\downarrow{}$& - & Image-level & Pixel-level & Image-level & Pixel-level \\
\midrule
RD~\cite{deng2022anomaly} & CVPR2022 & 95.8/97.8/95.0 & 95.1/51.5/90.7 & 88.4/89.9/86.7 & 96.8/38.7/87.0 \\
UniAD~\cite{you2022unified} & NeurIPS2022 & 95.0/97.7/94.5 & 95.8/46.8/90.0 & 90.7/92.2/86.9 & 98.3/38.1/88.8 \\
ViTAD~\cite{joy2025vitad} & CVIU2025 & 97.7/98.9/96.5 & 97.1/57.0/90.9 & 89.1/90.4/85.4 & 98.0/39.8/84.2 \\
MambaAD~\cite{he2024mambaad} & NeurIPS2024 & 94.7/97.7/94.5 & 96.3/51.5/90.5 & 90.2/91.4/87.5 & 97.7/39.4/87.9 \\
INP-Former~\cite{luo2025exploring} & CVPR2025 & 99.2/99.5/98.6 &  \textbf{98.2/60.7/93.8} & 95.2/95.7/91.9 & 98.8/44.4/\textbf{91.5} \\
CCL~\cite{fan2025salvaging} & ICCV2025 & 98.2/99.2/97.0 & 97.3/57.1/92.6 & 93.6/94.3/90.3 & 98.4/45.4/91.1 \\
\midrule
UniMMAD & Ours & \textbf{99.4/99.6/98.7} & 98.1/60.2/93.0 &  \textbf{95.5/96.2/92.4} &  \textbf{98.9/47.2}/91.3 \\
\bottomrule
\end{tabular}
  \end{adjustbox}
  \vspace{-2mm}
\end{table*}
\section{More Quantitative Comparison}
\label{sec:More Quantitative Comparison}
In this section, we compare additional metrics and comparison methods. As shown in Tab.~\ref{tab:appendix_mvtec_visa}, we compare with the latest method, CCL~\cite{fan2025salvaging}, using the same WideResNet50 backbone, and our approach achieves overall superior results.

\section{Ablation on  Number of Activated  Experts}
\label{sec:Ablation_exp_num}
\begin{figure}[htbp]
  \centering
    \includegraphics[width=1.0\linewidth]{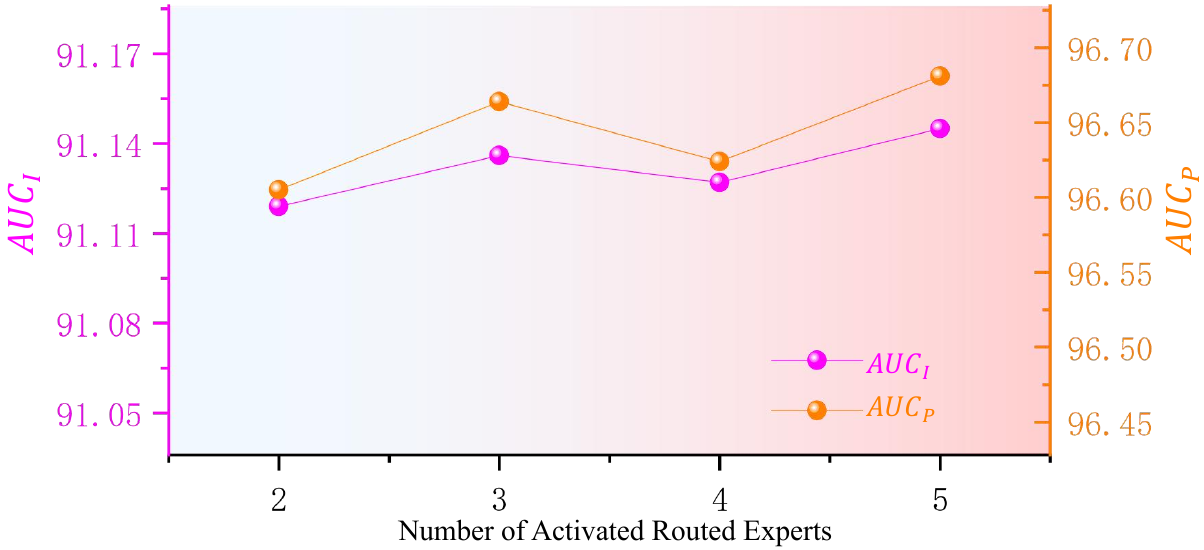}
  \caption{  Impact of the number of activated routed experts on performance.}
  \label{fig:topk_ablation}
\end{figure}

We conduct an ablation study on the number of activated routed experts.	
As shown in Fig.~\ref{fig:topk_ablation}, as the number of activated experts increases, the model's image and pixel-level metrics show a slight improvement.	
However, each additional expert also increases the number of dynamic filtering operations, reducing inference efficiency.	
To balance inference efficiency and performance, we set the number of activated experts to 2.	
This is a commonly used activation number, adopted by many MoE-based methods~\cite{lepikhin2020gshard,fedus2022switch}.	

\begin{table*}[!tbp]
  \caption{	Overview of the Datasets.}
  \label{tab:datasets_list}
  \centering
  \begin{adjustbox}{width=0.85\linewidth}
  \begin{tabular}{l|l|l|l|c|cc|cc}
\toprule
\multicolumn{2}{c|}{\multirow{2}{*}{Datasets}}&\multirow{2}{*}{Fields}&\multirow{2}{*}{Modality} & \multirow{2}{*}{\#Classes} &  \multicolumn{2}{c|}{Train} &  \multicolumn{2}{c}{Test} \\
% \cmidrule{5-6}
% \cmidrule{7-8}
 \multicolumn{2}{c|}{}&  & & & \#Items & \#Images & \#Items & \#Images \\
\midrule
\multicolumn{2}{l|}{MVTec-3D~\cite{bergmann2021mvtec}} & Industrial & RGB, Depth & 10 & 2656 & 5312 & 1197 & 2394 \\
\multicolumn{2}{l|}{Eyecandies~\cite{bonfiglioli2022eyecandies}} & Synthetic & Syn\_RGB, Syn\_Depth & 10 & 8751 & 17502 & 500 & 1000 \\
\multicolumn{2}{l|}{MulSen-AD~\cite{li2024multi}} & Industrial & Infrared & 15 & 1391 & 1391 & 644 & 644 \\

\multicolumn{2}{l|}{MVTec-AD~\cite{bergmann2019mvtec}} & Industrial & RGB & 15 & 3629 & 3629 & 2180 & 2180 \\ 
\multicolumn{2}{l|}{VisA~\cite{zou2022spot}} & Industrial & RGB & 12 & 8659 & 8659 & 2162 & 2162
\\ \multicolumn{2}{l|}{Brats2020~\cite{brats2020}} & Medical & Flair, T1, T2, T1CE& 1 & 1183 & 4732 & 167 & 668 \\
\midrule
\multirow{3}[4]{*}[0.5em]{\rotatebox{90}{UniMed}}&{Hyper-Kvasir~\cite{tian2021constrained}} & Medical & Endoscopy & 1 & 2020 & 2020 & 184 & 184 \\
&{Retinal OCT~\cite{hu2019automated}} & Medical & OCT & 1 & 1009 & 1009 & 270 & 270 \\
&{Liver CT~\cite{bao2024bmad}} & Medical & CT & 1 & 404 & 404 & 158 & 158 \\
\bottomrule
\end{tabular}
  \end{adjustbox}
\end{table*}

\begin{figure}[htbp]
  \centering
    \includegraphics[width=1.0\linewidth]{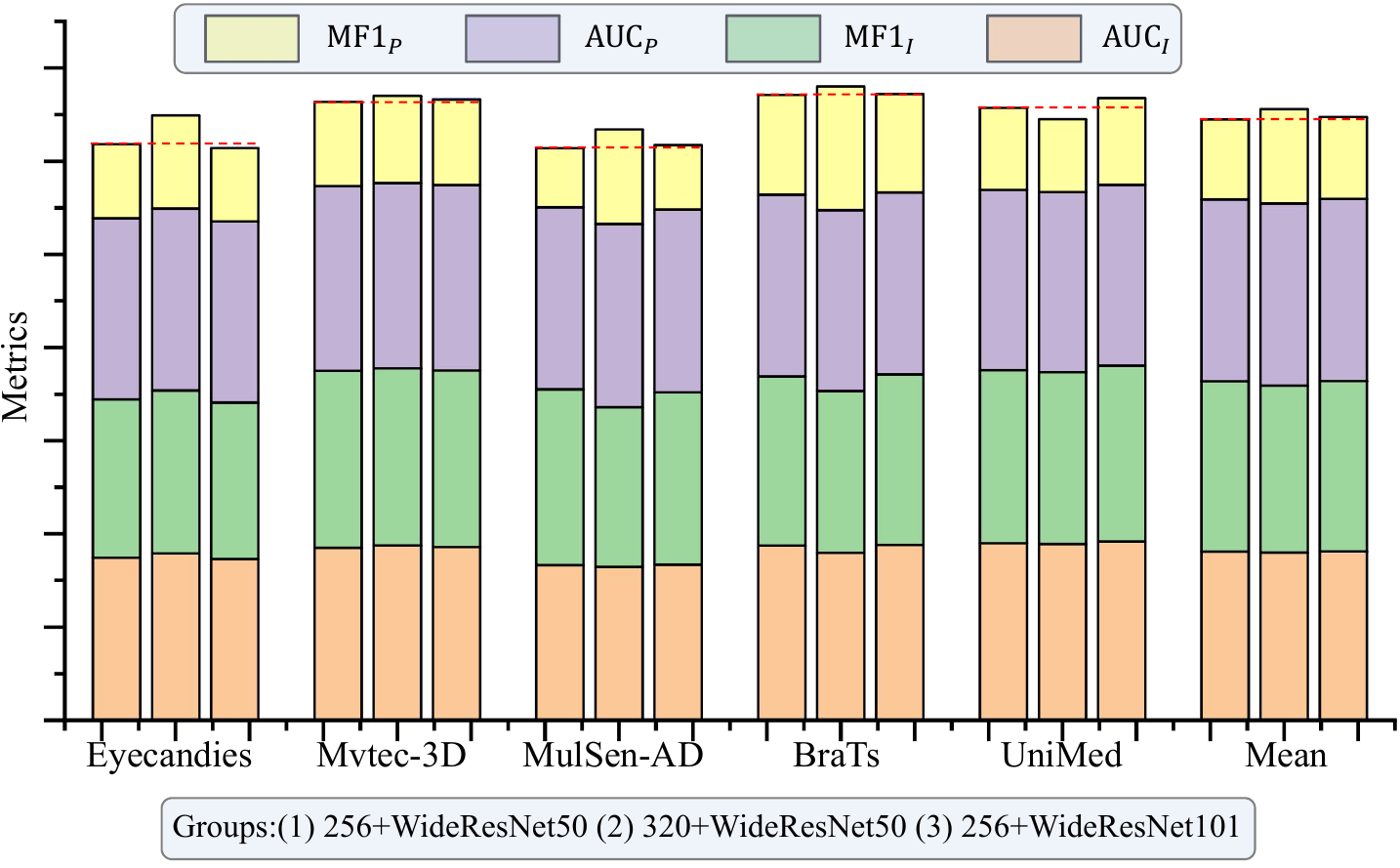}
  \caption{  Impact of input resolution and backbone. Each group is divided into  three settings: (1) 256$\times$256 resolution + WideResNet50, (2) 320$\times$320 resolution + WideResNet50, and (3) 256$\times$256 resolution + WideResNet101.	 }
  \label{fig:imgsize_backbone}
\end{figure}

\begin{table*}[!htbp]
  \caption{	 Per-Class comparison on the BraTs2020~\cite{brats2020} Dataset.}
  \label{tab:appendix_brats}
  \centering
  \begin{adjustbox}{width=0.85\linewidth}
  
\begin{tabular}{c|c|ccc|ccc}
\toprule
Methods&Publication& \zhaoyuan{$\text{AUC}_I$(\%)}& \zhaoyuan{$\text{AP}_I$(\%)} & \zhaoyuan{$\text{MF1}_I$ (\%)} & \zhaoyuan{$\text{AUC}_P$ (\%)}  & \zhaoyuan{$\text{MF1}_P$ (\%)} & \zhaoyuan{$\text{AUPRO}$ (\%)}\\
\midrule
\multicolumn{8}{c}{\textbf{Specialized Models (One model for one dataset)}} \\
\midrule
PatchCore~\cite{roth2022towards}+MMRD~\cite{gu2024rethinking} & - & 91.8 & 95.7 & 90.5 & 95.7 & 46.7 & 82.6 \\
\midrule
\multicolumn{8}{c}{\textbf{Generalist Models (One model performs all AD tasks without fine-tuning)}} \\
\midrule
AdaCLIP~\cite{cao2024adaclip} & ECCV'24 & 70.7 & 81.7 & 81.6 & 96.6 & 48.4 & 78.2 \\
MVFA~\cite{huang2024adapting} & CVPR'24 & 63.7 & 78.7 & 79.9 & 94.3 & 33.4 & 67.5 \\
AA-CLIP~\cite{ma2025aa} & CVPR'25 & 57.6 & 69.0 & 80.2 & 95.0 & 30.0 & 77.2 \\
\midrule
\multicolumn{8}{c}{ \textbf{Unified  Multi-modal and Multi-class Model}} \\
\midrule

\rowcolor{blue!10}UniMMAD & Ours  & \textbf{95.8} & \textbf{98.0} & \textbf{91.1} & \textbf{97.5} & \textbf{51.4} & \textbf{84.4} \\
\bottomrule
\end{tabular}
  \end{adjustbox}
\end{table*}

\begin{table*}[!htbp]
  \caption{	
  Per-Class comparison on the MVTec-AD~\cite{bergmann2019mvtec} Dataset for image-level metrics: $\text{AUC}_I $/$\text{AP}_I $ / $ \text{MF1}_I$. }
  \label{tab:mvtecAd_image}
  \centering
  \begin{adjustbox}{width=0.85\linewidth}
  \begin{tabular}{c|ccccc|>{\columncolor{blue!10}}c}
\toprule
Method$\xrightarrow{}$ &RD~\cite{deng2022anomaly} & UniAD~\cite{you2022unified} & ViTAD~\cite{joy2025vitad} & MambaAD~\cite{he2024mambaad} & INP-Former~\cite{luo2025exploring} & UniMMAD \\
Class$\downarrow$ & CVPR'22 & NeurIPS'22 & CVIU'25 & NeurIPS'24 & CVPR'25 & Ours \\
\midrule
Capsule & 91.4/98.0/93.9 & 85.8/96.6/91.9 & 94.0/98.6/95.4 & 84.7/96.2/93.3 & 97.4/98.2/96.7 & \textbf{97.6}/\textbf{99.3}/\textbf{97.2} \\

Carpet & 97.9/99.4/96.0 & 99.9/99.9/99.4 & 99.4/99.8/99.4 & 96.7/99.0/95.4 & \textbf{100.}/\textbf{100.}/\textbf{100.} & 99.3/99.7/98.3 \\

Grid & 92.3/97.2/93.2 & 97.2/99.1/96.4 & 99.3/99.7/99.1 & 96.8/99.0/97.2 & \textbf{100.}/\textbf{100.}/\textbf{100.} & \textbf{100.}/99.7/\textbf{100.} \\

Leather & \textbf{100.}/\textbf{100.}/\textbf{100.} & \textbf{100.}/\textbf{100.}/\textbf{100.} & \textbf{100.}/\textbf{100.}/\textbf{100.} & \textbf{100.}/\textbf{100.}/\textbf{100.} & \textbf{100.}/\textbf{100.}/\textbf{100.} & \textbf{100.}/99.8/\textbf{100.} \\

Tile & 98.5/99.4/96.4 & 99.1/99.6/97.0 & \textbf{100.}/\textbf{100.}/\textbf{100.} & 99.4/99.7/97.6 & \textbf{100.}/\textbf{100.}/\textbf{100.} & \textbf{100.}/99.8/\textbf{100.} \\

Wood & 99.2/99.7/\textbf{98.3} & 98.2/99.4/97.5 & 98.6/99.5/96.7 & \textbf{99.5}/\textbf{99.8}/\textbf{98.3} & 99.2/99.1/98.1 & 98.8/99.4/97.6 \\

Bottle & 99.7/99.9/98.4 & 99.9/99.9/99.2 & \textbf{100.}/\textbf{100.}/\textbf{100.} & 99.4/99.8/98.4 & \textbf{100.}/\textbf{100.}/\textbf{100.} & \textbf{100.}/99.8/\textbf{100.} \\

Cable & 78.9/86.8/80.5 & 84.5/91.9/81.2 & 97.3/98.4/93.3 & 85.4/91.2/83.7 & 99.5/\textbf{99.9}/\textbf{98.4} & \textbf{99.6}/99.5/97.3 \\

Hazelnut & 99.8/99.9/98.5 & 99.9/99.9/99.2 & 99.1/99.5/96.4 & 99.9/99.9/99.2 & \textbf{100.}/\textbf{100.}/\textbf{100.} & \textbf{100.}/99.7/\textbf{100.} \\

Metal nut & 99.1/99.7/97.3 & 99.3/99.8/97.8 & 99.3/99.8/98.4 & 97.5/99.4/95.7 & \textbf{100.}/\textbf{100.}/\textbf{100.} & \textbf{100.}/99.8/\textbf{100.} \\

Pill & 95.7/99.2/95.7 & 94.4/98.9/94.5 & 96.9/99.4/96.1 & 88.2/97.6/93.1 & \textbf{98.3}/99.1/96.8 & 97.9/\textbf{99.6}/\textbf{97.1} \\

Screw & 96.1/98.7/94.9 & 83.1/93.0/88.0 & 87.1/95.5/88.2 & 89.0/96.0/90.1 & 95.0/98.2/94.0 & \textbf{98.2}/\textbf{99.4}/\textbf{97.0} \\

Toothbrush & 99.1/99.6/98.3 & 91.9/96.4/93.7 & 99.7/99.8/98.3 & 97.7/99.1/95.2 & \textbf{100.}/\textbf{100.}/\textbf{100.} & 99.4/99.3/98.3 \\

Transistor & 88.8/88.3/85.0 & 93.7/91.0/84.4 & 95.9/94.2/88.6 & 89.2/88.8/83.3 & 99.0/98.3/95.3 & \textbf{99.6}/\textbf{98.7}/\textbf{98.7} \\

Zipper & 99.3/99.8/98.7 & 97.6/99.3/97.1 & 97.6/99.2/97.5 & 96.5/98.9/96.7 & \textbf{100.}/\textbf{100.}/\textbf{100.} & 99.6/99.8/99.1 \\

\midrule
Mean & 95.7/97.7/95.0 & 95.0/97.7/94.5 & 97.7/98.9/96.5 & 94.7/97.6/94.5 & 99.2/99.5/98.6 & \textbf{99.4}/\textbf{99.6}/\textbf{98.7} \\

\bottomrule
\end{tabular}

  \end{adjustbox}
\end{table*}
\begin{table*}[!htbp]
 \caption{	
  Per-Class comparison on the MVTec-AD~\cite{bergmann2019mvtec} dataset for pixel-level metrics: $\text{AUC}_P $/$\text{MF1}_P $ / $ \text{AUPRO}$. }
  \label{tab:mvtecAd_pixel}
  \centering
  \begin{adjustbox}{width=0.9\linewidth}
  \begin{tabular}{c|ccccc|>{\columncolor{blue!10}}c}
\toprule
Method$\xrightarrow{}$ &RD~\cite{deng2022anomaly} & UniAD~\cite{you2022unified} & ViTAD~\cite{joy2025vitad} & MambaAD~\cite{he2024mambaad} & INP-Former~\cite{luo2025exploring} & UniMMAD \\
Class$\downarrow$ & CVPR'22 & NeurIPS'22 & CVIU'25 & NeurIPS'24 & CVPR'25 & Ours \\
\midrule
Capsule & 98.5/47.5/95.0 & 98.5/48.3/92.5 & 97.9/47.9/92.2 & 98.6/44.4/94.3 & \textbf{99.0}/\textbf{52.1}/\textbf{95.6} & 98.9/50.2/92.5 \\

Carpet & 98.9/60.2/95.4 & 98.5/56.1/\textbf{95.6} & 98.8/\textbf{62.2}/94.2 & 98.5/56.4/93.8 & \textbf{99.1}/60.8/95.3 & 99.0/\textbf{62.2}/95.1 \\

Grid & 98.8/43.3/\textbf{96.1} & 96.2/28.4/90.8 & 98.4/36.5/95.3 & 98.8/46.8/96.0 & 98.4/40.4/94.6 & \textbf{99.0}/\textbf{47.2}/95.5 \\

Leather & 99.3/47.0/97.7 & 98.9/43.5/\textbf{98.1} & \textbf{99.5}/\textbf{53.5}/97.9 & 99.1/43.7/97.6 & 99.2/42.9/96.1 & 99.4/52.8/97.4 \\

Tile & 94.4/57.1/84.0 & 92.4/51.6/81.1 & 96.4/\textbf{68.1}/87.4 & 92.5/53.5/79.0 & \textbf{97.5}/67.5/\textbf{91.0} & 96.0/63.7/89.0 \\

Wood & 95.2/50.6/91.0 & 92.8/43.1/89.6 & 95.9/56.0/87.6 & 94.3/44.7/89.6 & \textbf{96.4}/\textbf{59.5}/\textbf{91.3} & 94.9/50.7/88.0 \\

Bottle & 97.3/66.2/93.5 & 97.4/64.9/92.7 & 98.7/73.9/94.6 & 97.2/64.8/93.2 & \textbf{99.0}/\textbf{78.6}/\textbf{95.6} & 98.7/74.4/93.4 \\

Cable & 77.4/30.8/73.5 & 91.4/32.1/77.0 & 92.4/41.4/85.4 & 88.2/33.1/78.9 & \textbf{98.7}/\textbf{67.5}/\textbf{94.4} & 97.8/60.8/91.9 \\

Hazelnut & 98.5/59.1/\textbf{95.6} & 97.9/56.0/94.6 & 98.7/61.3/94.7 & 98.5/59.3/95.3 & \textbf{99.2}/\textbf{68.7}/95.3 & 98.9/64.0/94.8 \\

Metal nut & 91.3/57.1/89.1 & 92.2/57.5/87.2 & 95.5/74.7/92.1 & 93.9/62.7/90.4 & 96.6/79.2/93.3 & \textbf{97.4}/\textbf{81.8}/\textbf{94.1} \\

Pill & 96.6/58.7/95.6 & 94.5/45.5/95.4 & \textbf{98.6}/\textbf{75.4}/95.6 & 97.4/64.0/95.4 & 97.2/65.1/\textbf{96.5} & 98.1/69.3/95.5 \\

Screw & 99.0/41.0/95.8 & 98.6/33.9/93.7 & 98.6/38.7/92.6 & 99.0/44.4/95.7 & \textbf{99.4}/\textbf{46.9}/\textbf{96.0} & \textbf{99.4}/45.7/\textbf{96.0} \\

Toothbrush & 98.7/55.7/90.8 & 97.9/45.0/85.1 & \textbf{99.1}/\textbf{64.9}/90.9 & 98.8/57.6/89.6 & 99.0/58.8/\textbf{93.5} & \textbf{99.1}/62.4/91.7 \\

Transistor & 83.4/39.5/71.0 & 92.1/45.4/82.8 & 92.0/51.4/74.0 & 90.3/41.5/73.4 & \textbf{96.0}/\textbf{61.9}/84.2 & 95.7/61.2/\textbf{85.5} \\

Zipper & \textbf{98.2}/58.1/\textbf{94.7} & 97.1/49.3/92.4 & 95.8/49.8/89.4 & 98.0/55.5/94.0 & 98.1/\textbf{59.6}/94.2 & 97.8/55.5/93.2 \\

\midrule
Mean &95.0/51.5/90.6 & 95.8/46.7/89.9 & 97.1/57.0/90.9 & 96.2/51.5/90.4 & \textbf{98.2}/\textbf{60.6}/\textbf{93.8} & 98.1/60.2/93.0 \\

\bottomrule
\end{tabular}

  \end{adjustbox}
\end{table*}
 
\section{Influence of  Resolution and Backbone}  
\label{sec:resolution_backbone}
As shown in Fig~.\ref{fig:imgsize_backbone}, we conducted an ablation study to assess the impact of input resolution and backbone architecture on model performance.	
Overall, larger image sizes and more powerful pre-trained models resulted in improved performance, demonstrating the scalability of our model in these aspects.
This highlights the model's scalability in both input resolution and backbone architecture.	
Specifically, larger image sizes result in more significant improvements for the RGB-D datasets Eyecandies and MVTec-3D, likely due to the correlation with anomaly size and point cloud accuracy.
Higher resolutions help the model capture finer defect localization.	
Larger backbones improve anomaly localization for medical lesions, likely due to the enhanced generalization ability of larger models in medical image tasks.

\section{Per-Class  Results}
\label{sec:Per-Class Results}
This section presents a class-wise comparison between the proposed method and recent state-of-the-art baselines.	
Tabs.~\ref{tab:mvtecAd_image}–\ref{tab:visa_pixel} illustrate that our method outperforms mainstream multi-class approaches on the widely used datasets MVTec-AD and VisA.
Tab.~\ref{tab:mvtec3d_image} and Tab.\ref{tab:mvtec3d_pixel} summarize image-level and pixel-level results on MVTec-3D \cite{bergmann2021mvtec}.	 The proposed approach attains the best scores on categories dominated by geometric defects (e.g., Potato) and on anomalies that benefit from multi-modal cues (e.g., Carrot).	 
On the synthetic Eyecandies benchmark (Tabs.~\ref{tab:eyecandies_image}–\ref{tab:eyecandies_pixel}), our model achieves high accuracy on the ChocolateCookie and PeppermintCandy classes.
These results indicate that the method effectively exploits complementary RGB–Depth information to pinpoint fine-grained anomalies.	
Tab.~\ref{tab:mulsen_image} and \ref{tab:mulsen_pixel} report infrared-based results on MulSen-AD, highlighting the model’s strength in detecting subsurface defects.	
In addition, Tab.~\ref{tab:UniMed} and \ref{tab:appendix_brats} list results on UniMed and BraTS 2020 datasets, showing that the unified model remains competitive despite large domain shifts.	
Overall, the class-wise analysis confirms that our unified framework generalizes across modalities and defect types, consistently outperforming prior work.

\section{More Qualitative Comparison}
\label{sec:More Qualitative Comparison}
\label{sec:qualitative}
\begin{figure*}[!htbp]
  \centering
    \includegraphics[width=1.0\linewidth]{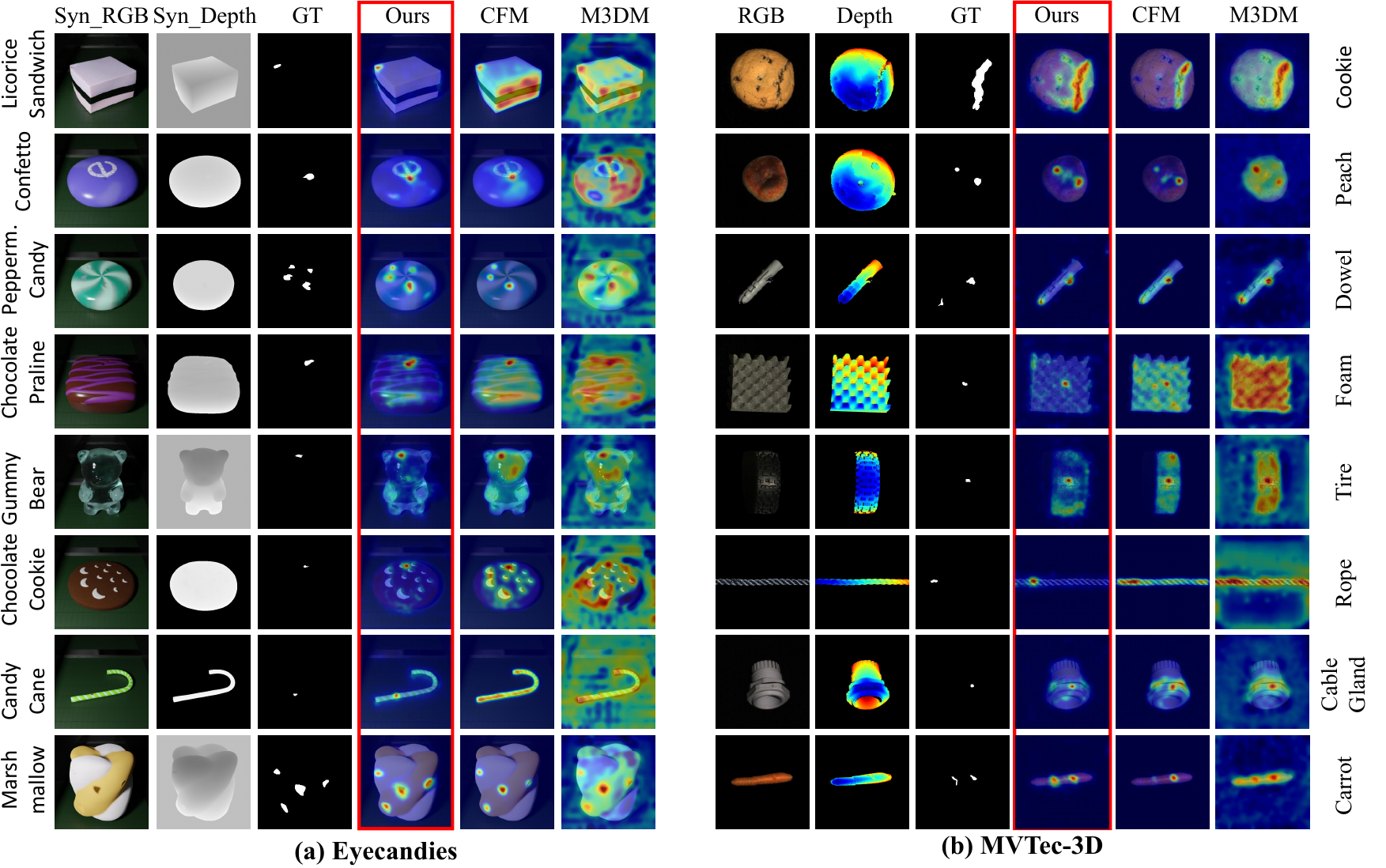}
  \caption{  Qualitative comparison on MVTec-3D and Eyecandies, with our method highlighted in the red box.}
  \label{fig:appendix_mvtec3d_eyecabdies_quali}
\end{figure*}

\begin{figure*}[!htbp]
  \centering
    \includegraphics[width=0.85\linewidth]{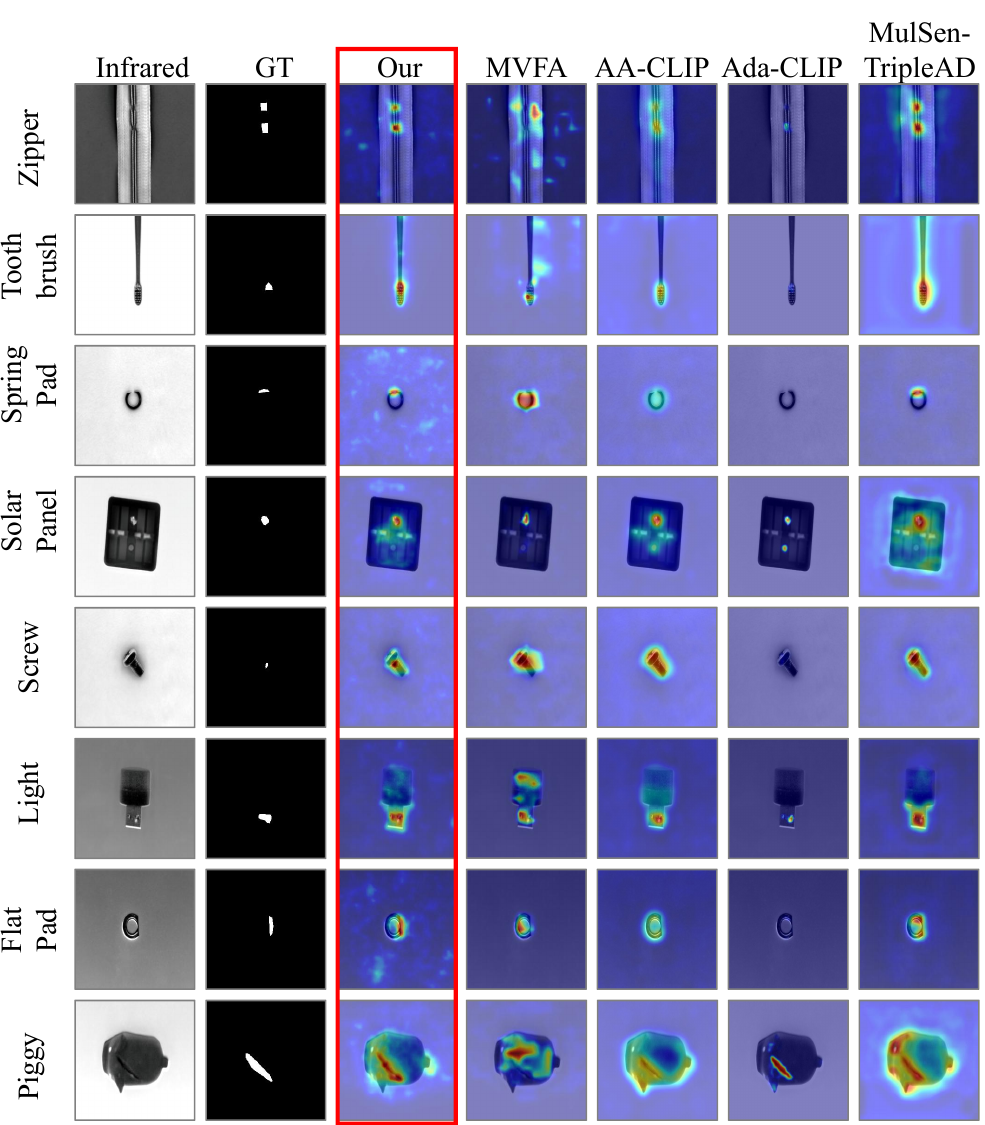}
  \caption{  Qualitative comparison on MulSen-AD, with our method highlighted in the red box.}
  \label{fig:appendix_mulsenad_quali}
\end{figure*}

\begin{figure*}[!htbp]
  \centering
    \includegraphics[width=1.0\linewidth]{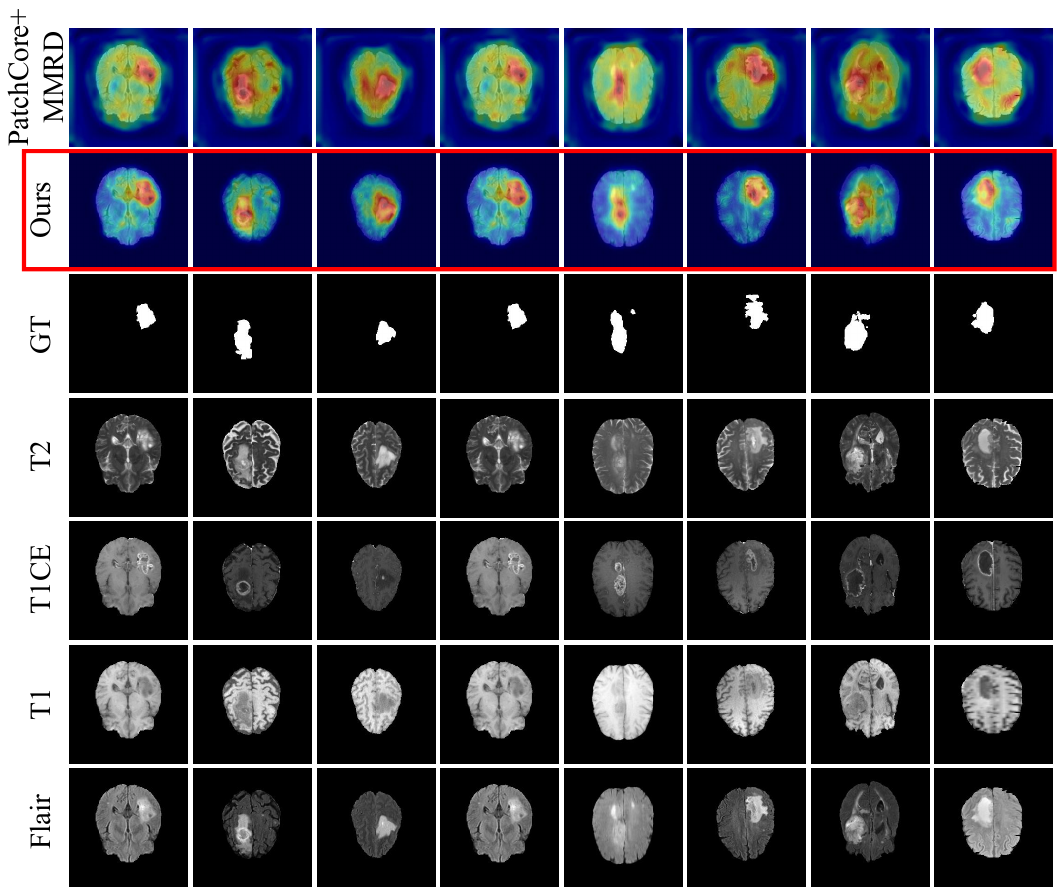}
  \caption{  Qualitative comparison on BraTs2020, with our method highlighted in the red box.}
  \label{fig:appendix_brain_quali}
\end{figure*}

\begin{figure*}[htbp]
  \centering
    \includegraphics[width=1.0\linewidth]{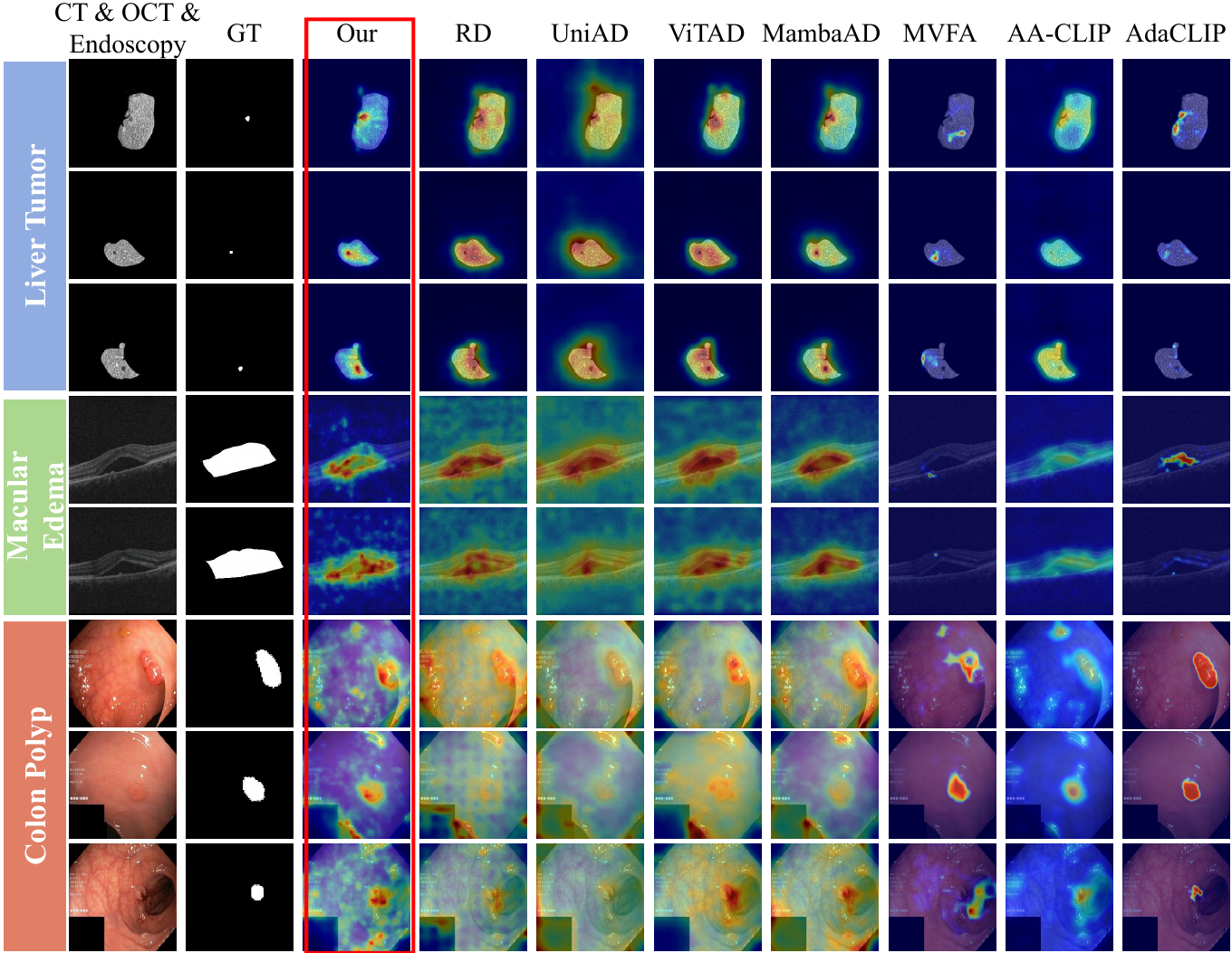}
  \caption{  Qualitative comparison on UniMed, with our method highlighted in the red box.}
  \label{fig:appendix_unimed_quali}
\end{figure*}

Fig.~\ref{fig:appendix_mvtec3d_eyecabdies_quali}, Fig.~\ref{fig:appendix_brain_quali}, Fig.~\ref{fig:appendix_unimed_quali}, and Fig.~\ref{fig:appendix_mulsenad_quali} show more qualitative comparison on MVTec-3D (industrial  RGB-D), Eyecandies (synthetic RGB-D), BraTs2020 (Brain Tumor), UniMed  (three medical  datasets) and MulSen-AD (industrial infrared). It demonstrates proposed method can accurately segment defect and lesion for a wide range of classes. 
In RGB-D datasets such as Cookie, our heatmaps highlight elongated defects along fiber directions. The model also emphasizes depth-based holes (e.g., Peach) and RGB-based color contamination (e.g., Candy Cane).	
In medical datasets such as UniMed, our method effectively separates boundary-blurred lesions from normal tissues, enabling high-confidence localization of conditions like colon polyps and brain tumors.	
Some generalist models like MVFA and AdaCLIP also show confident segmentation for colon polyps, possibly due to overlaps with their supervised training data.	
However, they incorrectly focus on normal regions in MulSen-AD, Liver Tumor, and Macular Edema.	
In contrast, UniMMAD consistently delivers accurate defect localization across diverse scenes, modalities, and object categories.	
These results highlight the effectiveness of the proposed ``general  $\rightarrow$ specific'' paradigm and domain-specific reconstruction pathways in C-MoE, which mitigate cross-domain reconstruction interference.
UniMMAD demonstrates precise defect localization and sharper boundaries.

\begin{table*}[!htbp]
  \caption{	
  Per-Class comparison on the VisA~\cite{zou2022spot} dataset for image-level metrics: $\text{AUC}_I $/$\text{AP}_I $ / $ \text{MF1}_I$. }
  \label{tab:visa_image}
  \centering
  \begin{adjustbox}{width=0.9\linewidth}
  \begin{tabular}{c|ccccc|>{\columncolor{blue!10}}c}
\toprule
Method$\xrightarrow{}$ &RD~\cite{deng2022anomaly} & UniAD~\cite{you2022unified} & ViTAD~\cite{joy2025vitad} & MambaAD~\cite{he2024mambaad} & INP-Former~\cite{luo2025exploring} & UniMMAD \\
Class$\downarrow$ & CVPR'22 & NeurIPS'22 & CVIU'25 & NeurIPS'24 & CVPR'25 & Ours \\
\midrule
Candle & 89.0/89.3/82.2 & 93.8/94.4/86.2 & 87.3/88.2/82.1 & 90.3/91.2/84.2 & \textbf{97.4}/\textbf{97.2}/\textbf{92.6} & 95.3/94.8/90.3 \\

Capsules & 75.2/86.0/78.4 & 74.8/85.2/78.6 & 78.6/87.0/78.1 & 76.8/84.1/80.8 & \textbf{90.3}/\textbf{94.8}/\textbf{88.5} & 88.5/93.6/84.6 \\

Cashew & 87.8/93.6/88.1 & 91.1/95.6/88.6 & 82.2/91.1/86.0 & 88.5/94.2/88.1 & 95.5/97.6/\textbf{93.9} & \textbf{95.7}/\textbf{98.0}/93.2 \\

Chewinggum & 91.1/95.5/88.8 & 97.4/98.8/95.7 & 92.1/96.3/88.3 & 89.8/95.1/87.0 & 98.3/99.3/96.4 & \textbf{98.8}/\textbf{99.5}/\textbf{97.9} \\

Fryum & 94.7/97.6/90.8 & 92.3/96.3/91.2 & 93.3/97.0/89.6 & 94.8/97.8/92.0 & \textbf{98.0}/\textbf{99.0}/\textbf{94.8} & 96.2/98.3/92.8 \\

Macaroni1 & 92.2/89.9/86.3 & 87.5/83.1/79.6 & 84.5/81.2/77.9 & 88.9/87.7/82.2 & 94.2/93.3/88.6 & \textbf{97.2}/\textbf{97.2}/\textbf{91.5} \\

Macaroni2 & 80.7/76.6/75.7 & 76.2/76.0/69.2 & 78.3/71.8/73.9 & 69.6/60.9/71.0 & 82.5/79.8/78.5 & \textbf{83.3}/\textbf{82.4}/\textbf{79.4} \\

Pcb1 & 95.1/94.9/91.0 & 96.5/96.0/92.7 & 95.3/93.7/90.4 & 96.1/95.9/90.6 & \textbf{97.3}/\textbf{96.7}/\textbf{95.1} & 97.1/96.5/95.0 \\

Pcb2 & \textbf{97.4}/\textbf{97.6}/93.1 & 92.4/92.8/86.2 & 90.8/89.4/85.7 & 95.6/96.1/90.9 & 96.7/96.3/91.5 & 97.2/97.1/\textbf{94.0} \\

Pcb3 & 60.0/58.9/70.4 & 88.4/88.6/81.2 & 90.8/91.4/83.5 & 94.5/94.6/88.8 & 95.1/95.7/87.8 & \textbf{97.1}/\textbf{97.3}/\textbf{93.0} \\

Pcb4 & \textbf{99.9}/\textbf{99.9}/\textbf{99.0} & 99.4/99.4/96.4 & 98.9/98.7/96.1 & 99.7/99.7/97.5 & 99.7/99.8/97.5 & 99.8/99.6/98.0 \\

Pipe fryum & 97.3/98.6/95.5 & 97.9/99.1/96.0 & 95.9/98.0/92.8 & 97.7/98.7/96.5 & 97.3/98.8/97.5 & \textbf{99.8}/\textbf{99.7}/\textbf{99.0} \\

\midrule
Mean & 88.4/89.9/86.6 & 90.6/92.1/86.8 & 89.0/90.3/85.4 & 90.2/91.3/87.5 & 95.2/95.7/91.9 & \textbf{95.5}/\textbf{96.2}/\textbf{92.4} \\

\bottomrule
\end{tabular}

  \end{adjustbox}
\end{table*}
\begin{table*}[!htbp]
 \caption{	
  Per-Class comparison on the VisA~\cite{zou2022spot} dataset for pixel-level metrics: $\text{AUC}_P $/$\text{MF1}_P $ / $ \text{AUPRO}$. }
  \label{tab:visa_pixel}
  \centering
  \begin{adjustbox}{width=0.9\linewidth}
  \begin{tabular}{c|ccccc|>{\columncolor{blue!10}}c}
\toprule
Method$\xrightarrow{}$ &RD~\cite{deng2022anomaly} & UniAD~\cite{you2022unified} & ViTAD~\cite{joy2025vitad} & MambaAD~\cite{he2024mambaad} & INP-Former~\cite{luo2025exploring} & UniMMAD \\
Class$\downarrow$ & CVPR'22 & NeurIPS'22 & CVIU'25 & NeurIPS'24 & CVPR'25 & Ours \\
\midrule
Candle & 98.8/34.4/94.3 & 99.2/33.4/94.9 & 95.4/26.1/83.7 & 99.0/30.0/\textbf{95.9} & \textbf{99.4}/\textbf{37.9}/95.2 & 99.3/37.0/94.7 \\

Capsules & 99.0/56.2/90.9 & 97.4/40.6/77.6 & 98.1/41.9/74.7 & 98.4/46.2/88.0 & 99.4/56.5/91.4 & \textbf{99.6}/\textbf{65.8}/\textbf{91.9} \\

Cashew & 87.3/42.5/68.3 & 97.5/47.7/90.6 & 97.4/58.6/69.1 & 91.2/45.2/70.3 & 97.6/60.5/89.3 & \textbf{98.8}/\textbf{62.3}/\textbf{90.7} \\

Chewinggum & 97.3/52.7/68.8 & \textbf{99.3}/\textbf{61.9}/\textbf{87.3} & 97.2/57.3/68.8 & 96.6/52.9/62.0 & 98.6/58.8/82.9 & 98.4/59.9/81.9 \\

Fryum & 96.8/50.3/\textbf{92.6} & 96.9/50.4/86.5 & \textbf{97.3}/51.2/88.9 & 96.9/51.5/92.2 & 97.1/48.6/89.9 & 97.0/\textbf{52.5}/88.4 \\

Macaroni1 & \textbf{99.7}/32.2/\textbf{96.4} & 99.1/15.4/93.8 & 98.3/15.3/89.6 & 99.5/26.4/95.9 & 98.8/21.1/91.9 & 99.3/\textbf{32.8}/92.4 \\

Macaroni2 & \textbf{99.4}/\textbf{17.6}/\textbf{95.2} & 97.7/10.4/89.6 & 98.1/10.1/87.3 & 98.5/\;\;8.6/91.1 & 99.0/11.7/93.5 & 98.9/16.9/92.1 \\

Pcb1 & 99.2/54.0/95.1 & 99.2/57.0/91.2 & 99.3/56.8/88.7 & 99.5/65.8/\textbf{95.2} & \textbf{99.7}/68.5/94.9 & \textbf{99.7}/\textbf{72.9}/94.8 \\

Pcb2 & 97.7/29.4/90.9 & 98.1/19.4/86.1 & 98.0/21.1/83.4 & 98.5/24.8/\textbf{92.5} & \textbf{98.9}/\textbf{31.5}/89.9 & 98.8/26.5/\textbf{92.5} \\

Pcb3 & 89.7/\;\;3.5/67.7 & 98.3/28.2/86.7 & 98.0/28.5/88.0 & 98.5/29.5/\textbf{93.9} & \textbf{99.1}/\textbf{30.1}/91.7 & \textbf{99.1}/29.0/91.1 \\

Pcb4 & 97.3/34.6/87.3 & 97.5/39.6/86.1 & \textbf{98.9}/44.7/\textbf{92.7} & 96.1/35.8/81.7 & 98.7/\textbf{45.9}/92.1 & 98.4/43.1/90.9 \\

Pipe fryum & 98.9/56.7/\textbf{95.7} & 98.7/52.3/94.1 & \textbf{99.4}/66.0/94.3 & 99.0/55.2/95.2 & 99.2/60.9/94.9 & 99.3/\textbf{67.5}/94.2 \\

\midrule

Mean & 96.8/38.7/86.9 & 98.2/38.0/88.7 & 97.9/39.8/84.1 & 97.6/39.3/87.8 & 98.8/44.3/\textbf{91.5} & \textbf{98.9}/\textbf{47.2}/91.3 \\

\bottomrule
\end{tabular}

  \end{adjustbox}
\end{table*}
\begin{table*}[!htbp]
  \caption{	
  Per-Class comparison on the MVTec-3D~\cite{bergmann2021mvtec} dataset for image-level metrics: $\text{AUC}_I $/$\text{AP}_I $ / $ \text{MF1}_I$. }
  \label{tab:mvtec3d_image}
  \centering
  \begin{adjustbox}{width=0.9\linewidth}
  \begin{tabular}{c|cc|ccc|>{\columncolor{blue!10}}c}
\toprule
Method$\xrightarrow{}$ & M3DM~\cite{wang2023multimodal} & CFM~\cite{costanzino2024multimodal} & AdaCLIP~\cite{cao2024adaclip} & MVFA~\cite{huang2024adapting} & AACLIP~\cite{ma2025aa} & UniMMAD \\
Class$\downarrow$ & CVPR'23 & CVPR'24 & ECCV'24 & CVPR'24 & CVPR'25 & Ours \\
\midrule
Bagel & 98.7/\textbf{99.9}/\textbf{98.1} & \textbf{99.2}/99.8/97.8 & 85.3/95.6/92.1 & 62.1/87.4/89.3 & 78.5/93.8/89.2 & 94.4/98.5/94.2 \\

Cable gland & 84.7/96.1/92.3 & 81.8/95.1/89.8 & 61.6/87.8/90.2 & 57.7/82.6/89.2 & 49.5/82.3/89.7 & \textbf{94.8}/\textbf{98.5}/\textbf{96.0} \\

Carrot & 94.6/99.0/95.6 & 97.8/99.5/97.3 & 80.4/95.8/91.7 & 71.3/92.5/91.8 & 64.5/89.5/91.9 & \textbf{99.4}/\textbf{99.8}/\textbf{98.8} \\

Potato & \textbf{95.8}/\textbf{99.4}/\textbf{96.2} & 88.4/96.8/93.3 & 63.3/86.7/90.2 & 48.4/80.1/89.3 & 87.5/96.7/92.1 & 93.3/97.9/\textbf{96.2} \\

Rope & 91.6/96.8/89.1 & 93.8/97.5/90.9 & 82.5/90.9/85.5 & 91.6/96.6/91.5 & 81.3/91.8/84.1 & \textbf{97.5}/\textbf{99.0}/\textbf{97.1} \\

Foam & \textbf{91.4}/\textbf{98.0}/\textbf{94.3} & \textbf{91.4}/97.8/92.0 & 69.5/91.2/88.9 & 53.6/86.9/88.9 & 49.7/81.8/88.9 & 89.1/97.3/91.6 \\

Dowel & 89.2/95.7/92.3 & 92.7/98.2/92.4 & 73.1/92.7/89.7 & 55.4/84.0/88.9 & 46.3/79.5/87.9 & \textbf{95.9}/\textbf{98.8}/\textbf{96.2} \\

Tire & \textbf{87.3}/\textbf{96.0}/\textbf{92.1} & 86.6/95.8/90.4 & 61.6/85.8/87.4 & 46.9/76.8/87.4 & 74.4/88.2/90.5 & 70.4/88.0/88.7 \\

Peach & 92.3/97.8/95.1 & 93.4/98.2/95.0 & 77.4/92.7/92.0 & 67.0/86.9/90.0 & 87.6/96.1/91.2 & \textbf{95.8}/\textbf{98.8}/\textbf{96.3} \\

Cookie & 95.6/98.9/96.1 & 98.9/99.7/\textbf{98.5} & 86.4/95.5/91.8 & 73.7/89.9/88.4 & \textbf{99.1}/\textbf{99.8}/\textbf{98.5} & 94.2/98.3/93.4 \\

\midrule
Mean & 92.1/97.7/94.1 & 92.4/\textbf{97.8}/93.7 & 74.1/91.4/89.9 & 62.7/86.3/89.4 & 71.8/89.9/90.4 & \textbf{92.5}/97.5/\textbf{94.9} \\

\bottomrule
\end{tabular}

  \end{adjustbox}
\end{table*}
\begin{table*}[!htbp]
  \caption{	
  Per-Class comparison on the MVTec-3D~\cite{bergmann2021mvtec} dataset for pixel-level metrics: $\text{AUC}_P $/$\text{MF1}_P $ / $ \text{AUPRO}$. }
  \label{tab:mvtec3d_pixel}
  \centering
  \begin{adjustbox}{width=0.9\linewidth}
  \begin{tabular}{c|cc|ccc|>{\columncolor{blue!10}}c}
\toprule
Method$\xrightarrow{}$ & M3DM~\cite{wang2023multimodal} & CFM~\cite{costanzino2024multimodal} & AdaCLIP~\cite{cao2024adaclip} & MVFA~\cite{huang2024adapting} & AACLIP~\cite{ma2025aa} & UniMMAD \\
Class$\downarrow$ & CVPR'23 & CVPR'24 & ECCV'24 & CVPR'24 & CVPR'25 & Ours \\
\midrule
Bagel & \textbf{99.4}/51.4/96.4 & 99.3/\textbf{52.8}/\textbf{96.9} & 98.6/49.4/92.0 & 88.2/\;\;6.0/50.4 & 99.0/44.3/94.1 & 98.8/48.0/94.2 \\

Cable gland & \textbf{99.3}/\textbf{47.5}/97.0 & 98.0/28.2/93.1 & 96.1/30.6/85.6 & 91.7/\;\;4.0/72.9 & 96.1/29.6/86.7 & 99.2/44.6/\textbf{97.1} \\

Carrot & 99.6/41.4/97.7 & \textbf{99.8}/\textbf{53.7}/\textbf{97.9} & 98.7/27.6/95.2 & 97.3/13.6/91.3 & 99.6/44.1/97.6 & 99.6/43.6/\textbf{97.9} \\

Potato & 99.4/40.4/96.8 & 99.5/34.0/97.4 & 99.4/31.1/96.8 & 97.7/10.6/90.5 & \textbf{99.7}/\textbf{51.5}/\textbf{97.8} & 99.6/42.1/\textbf{97.8} \\

Rope & 99.6/49.4/\textbf{96.5} & \textbf{99.7}/\textbf{63.8}/96.3 & 98.6/46.1/93.4 & 97.3/29.1/87.7 & 99.5/53.3/95.4 & 99.4/39.5/96.4 \\

Foam & 98.5/43.5/93.7 & \textbf{98.7}/\textbf{48.1}/\textbf{94.1} & 90.7/42.7/70.3 & 91.9/28.0/73.0 & 95.1/\;\;9.9/84.6 & 97.4/38.9/90.3 \\

Dowel & 96.8/33.7/91.1 & 97.6/35.0/91.6 & 94.3/10.2/83.1 & 89.8/\;\;7.5/72.6 & 98.4/\textbf{37.1}/93.2 & \textbf{98.9}/35.1/\textbf{96.0} \\

Tire & \textbf{99.3}/\textbf{41.2}/\textbf{96.3} & 98.9/17.1/95.1 & 97.2/40.7/87.6 & 93.1/\;\;7.8/76.8 & 98.9/26.7/95.9 & 98.6/36.8/94.6 \\

Peach & \textbf{99.5}/42.4/97.4 & 99.2/43.0/96.4 & 98.3/31.1/93.3 & 95.1/\;\;8.8/81.6 & 97.6/\textbf{59.6}/94.7 & \textbf{99.5}/48.7/\textbf{97.5} \\

Cookie & 97.8/56.2/92.4 & 96.8/\textbf{65.5}/93.9 & 97.2/61.1/93.3 & 90.3/24.3/72.7 & 96.8/64.8/94.0 & \textbf{99.3}/63.8/\textbf{96.9} \\

\midrule
Mean & 98.9/\textbf{44.7}/95.5 & 98.7/44.1/95.3 & 96.9/37.1/89.1 & 93.2/14.0/76.9 & 98.0/42.1/93.4 & \textbf{99.0}/44.1/\textbf{95.9} \\

\bottomrule
\end{tabular}

  \end{adjustbox}
\end{table*}

\begin{table*}[!htbp]
  \caption{	
  Per-Class comparison on the Eyecandies~\cite{bonfiglioli2022eyecandies} dataset for image-level metrics: $\text{AUC}_I $/$\text{AP}_I $ / $ \text{MF1}_I$. }
  \label{tab:eyecandies_image}
  \centering
  \begin{adjustbox}{width=0.9\linewidth}
  \begin{tabular}{c|cc|ccc|>{\columncolor{blue!10}}c}
\toprule
Method$\xrightarrow{}$ & M3DM~\cite{wang2023multimodal} & CFM~\cite{costanzino2024multimodal} & AdaCLIP~\cite{cao2024adaclip} & MVFA~\cite{huang2024adapting} & AACLIP~\cite{ma2025aa} & UniMMAD \\
Class$\downarrow$ & CVPR'23 & CVPR'24 & ECCV'24 & CVPR'24 & CVPR'25 & Ours \\
\midrule
Gummybear & 75.2/81.3/73.2 & 79.3/\textbf{84.7}/77.3 & 58.2/62.2/66.7 & 55.0/50.4/66.7 & 52.2/55.5/67.6 & \textbf{90.7}/84.3/\textbf{89.7} \\

Lollipop & 78.7/61.7/69.0 & 80.7/\textbf{76.6}/71.0 & 70.0/52.6/59.1 & 63.6/43.8/56.2 & 42.5/26.5/49.2 & \textbf{85.1}/76.1/\textbf{78.7} \\

Marshmallow & 91.7/92.2/84.7 & 97.8/98.3/95.8 & 82.3/85.3/80.0 & 82.1/83.8/79.2 & 60.2/67.1/66.7 & \textbf{99.8}/\textbf{98.9}/\textbf{98.0} \\

Licoricesandwich & 79.8/80.1/80.0 & 88.6/91.8/84.3 & 63.4/58.9/69.3 & 69.0/65.3/69.6 & 47.5/49.4/67.6 & \textbf{90.0}/\textbf{92.6}/\textbf{86.9} \\

Chocolatepraline & 75.8/80.1/72.3 & \textbf{88.0}/\textbf{91.3}/82.6 & 86.6/80.0/\textbf{86.8} & 72.6/72.7/76.9 & 54.2/55.0/69.6 & 82.7/87.2/76.1 \\

Chocolatecookie & 74.4/70.6/74.1 & 88.2/90.9/83.7 & 78.7/85.3/76.2 & 58.1/61.6/69.3 & 55.8/55.6/66.7 & \textbf{97.9}/\textbf{98.1}/\textbf{96.0} \\

Peppermintcandy & \textbf{95.3}/\textbf{96.0}/\textbf{90.6} & 85.8/89.3/79.2 & 85.9/85.7/81.5 & 76.3/78.7/76.7 & 27.7/37.0/66.7 & 92.6/93.1/84.0 \\

Hazelnuttruffle & 53.4/55.8/66.7 & \textbf{71.7}/\textbf{80.1}/69.6 & 37.6/40.3/66.7 & 60.0/57.7/\textbf{70.0} & 42.6/42.4/67.6 & 61.9/64.6/69.6 \\

Confetto & 89.1/90.0/83.0 & 88.2/91.8/83.7 & 87.8/89.1/83.0 & 66.6/69.6/71.2 & 61.3/65.5/67.7 & \textbf{97.2}/\textbf{96.8}/\textbf{92.0} \\

Candycane & 60.3/62.2/66.7 & 49.6/47.5/68.6 & \textbf{80.6}/\textbf{77.0}/\textbf{79.3} & 44.8/50.7/66.7 & 37.8/43.4/66.7 & 57.4/57.2/69.5 \\

\midrule
Mean & 77.3/77.0/76.0 & 81.8/84.2/79.6 & 73.1/71.6/74.8 & 64.8/63.4/70.2 & 48.1/49.7/65.6 & \textbf{85.5}/\textbf{84.9}/\textbf{84.1} \\

\bottomrule
\end{tabular}

  \end{adjustbox}
\end{table*}
\begin{table*}[!htbp]
    \caption{	
  Per-Class comparison on Eyecandies~\cite{bonfiglioli2022eyecandies}  Dataset for pixel-level metrics: $\text{AUC}_P $/$\text{MF1}_P $ / $ \text{AUPRO}$. }
  \label{tab:eyecandies_pixel}
  \centering
  \begin{adjustbox}{width=0.9\linewidth}
  \begin{tabular}{c|cc|ccc|>{\columncolor{blue!10}}c}
\toprule
Method$\xrightarrow{}$ & M3DM~\cite{wang2023multimodal} & CFM~\cite{costanzino2024multimodal} & AdaCLIP~\cite{cao2024adaclip} & MVFA~\cite{huang2024adapting} & AACLIP~\cite{ma2025aa} & UniMMAD \\
Class$\downarrow$ & CVPR'23 & CVPR'24 & ECCV'24 & CVPR'24 & CVPR'25 & Ours \\
\midrule
Gummybear & 94.8/37.4/78.5 & 93.1/19.3/78.9 & \textbf{96.6}/25.2/87.6 & 88.6/\;\;4.7/53.7 & 88.3/\;\;6.6/64.8 & 95.9/\textbf{41.5}/\textbf{88.6} \\

Lollipop & 96.5/28.6/77.5 & \textbf{98.3}/22.2/\textbf{93.0} & 97.7/11.4/87.8 & 95.1/\;\;6.1/78.4 & 97.3/13.6/88.9 & 98.0/\textbf{30.2}/90.3 \\

Marshmallow & 98.9/64.5/93.7 & 99.2/\textbf{66.5}/93.0 & 97.8/56.7/89.1 & 90.8/23.5/62.2 & 95.3/30.7/81.4 & \textbf{99.5}/58.4/\textbf{96.6} \\

Licoricesandwich & 96.8/\textbf{43.4}/85.4 & 96.0/28.1/81.7 & 95.6/31.5/80.3 & 82.8/11.2/47.6 & 95.6/36.4/81.4 & \textbf{98.4}/39.4/\textbf{93.5} \\

Chocolatepraline & 95.6/\textbf{51.7}/78.4 & 93.3/46.1/71.4 & 86.0/46.2/78.3 & 76.4/19.5/49.0 & \textbf{98.1}/47.0/\textbf{91.9} & 93.2/44.0/83.0 \\

Chocolatecookie & 89.4/23.5/71.4 & 97.5/35.5/86.9 & 95.8/\textbf{46.3}/86.8 & 50.0/12.8/19.5 & 94.3/45.1/84.0 & \textbf{98.5}/43.5/\textbf{93.1} \\

Peppermintcandy & 94.3/44.3/84.7 & 95.8/19.0/86.0 & 97.3/\textbf{49.4}/91.1 & 85.1/12.7/56.0 & 94.3/33.8/61.6 & \textbf{99.4}/48.0/\textbf{97.2} \\

Hazelnuttruffle & 89.2/19.0/61.0 & \textbf{90.8}/\textbf{38.2}/61.6 & 87.6/\;\;7.1/53.5 & 71.6/\;\;4.3/42.1 & 87.8/\;\;5.6/\textbf{96.2} & 90.6/23.4/65.2 \\

Confetto & 95.6/45.9/85.2 & 97.5/37.3/90.1 & 98.9/33.1/96.1 & 90.2/13.0/70.0 & 99.1/43.5/93.1 & \textbf{99.6}/\textbf{56.1}/\textbf{98.0} \\

Candycane & 85.7/\;\;6.4/70.8 & 97.2/\;\;6.6/90.6 & \textbf{98.0}/\textbf{13.1}/\textbf{93.3} & 95.7/\;\;2.7/87.3 & 97.5/\;\;9.3/70.8 & 95.4/\;\;9.4/86.9 \\

\midrule
Mean & 93.7/36.5/78.7 & 95.8/31.9/83.3 & 95.1/32.0/84.4 & 82.6/11.1/56.6 & 94.7/27.2/81.4 & \textbf{96.9}/\textbf{39.4}/\textbf{89.2} \\

\bottomrule
\end{tabular}

  \end{adjustbox}
\end{table*}
\begin{table*}[!htbp]
  \caption{	Per-Class comparison across UniMed datasets: Reinal OCT / Liver CT / Hyper-Kvasir / Mean performance.}
  \label{tab:UniMed}
  \centering
  \begin{adjustbox}{width=1\linewidth}
  
\begin{tabular}{c|c|c|c|c|c|c}
\toprule
Methods& \textbf{$\text{AUC}_I$(\%)}& \textbf{$\text{AP}_I$(\%)} & \textbf{$\text{MF1}_I$ (\%)} & \textbf{$\text{AUC}_P$ (\%)}  & AUPRO& \textbf{$\text{MF1}_P$ (\%)} \\
\midrule
\multicolumn{7}{c}{\textbf{Specialized Models (One model for one dataset)}} \\
\midrule
UniAD~\cite{you2022unified} & 90.4/83.7/98.4/90.8 & 95.9/90.8/99.9/95.5 & 91.0/83.6/98.1/90.9 & 95.1/97.2/68.6/87.0 & 81.5/90.4/32.9/68.2 & 63.3/30.1/24.9/39.4 \\
RD~\cite{deng2022anomaly} & 93.3/90.7/96.3/93.4 & 97.3/94.6/97.0/96.3 & 91.5/90.3/92.9/91.6 & 96.2/96.3/79.0/90.5 & 86.1/91.1/46.5/74.5 & 66.0/23.3/30.7/40.0 \\
ViTAD~\cite{joy2025vitad} & 95.3/92.9/98.3/95.5 & 98.1/95.5/98.7/97.4 & 93.8/92.6/94.4/93.6 & 94.5/97.4/83.5/91.8 & 80.3/93.4/52.6/75.4 & 63.4/35.1/34.9/44.4 \\
MambaAD~\cite{he2024mambaad} & 96.2/94.6/97.1/96.0 & 98.6/97.4/98.4/98.1 & 94.1/91.9/94.7/93.6 & 96.6/96.5/82.0/91.7 & 86.7/91.5/55.8/78.0 & 68.4/25.0/33.8/42.4 \\
SimpleNet~\cite{liu2023simplenet} & 90.4/80.2/97.6/89.4 & 96.4/91.3/97.4/95.0 & 88.7/80.5/96.3/88.5 & 92.0/92.5/76.1/86.9 & 72.3/64.3/36.7/57.8 & 55.8/34.4/27.2/39.1 \\
INP-Former~\cite{luo2025exploring} &97.4/91.0/99.8/96.1 & 99.0/95.5/99.1/97.9 & 94.9/90.1/98.1/94.4 & 95.2/97.2/85.6/92.7 & 81.6/90.3/57.5/76.5 & 67.2/33.2/36.0/45.5 \\

\midrule
\multicolumn{7}{c}{\textbf{Generalist Models (One model performs all AD tasks without fine-tuning)}} \\
\midrule
AdaCLIP~\cite{cao2024adaclip} & 82.4/70.8/98.2/83.8 & 92.6/82.2/98.8/91.2 & 85.1/84.8/95.9/88.6 & 91.2/92.2/88.8/90.7 & 78.2/87.7/67.5/77.8 & 51.9/18.0/58.7/42.9 \\

MVFA~\cite{huang2024adapting} & 88.6/89.2/88.5/88.8 & 94.7/95.1/91.7/93.8 & 91.0/87.3/83.3/87.2 & 89.2/96.3/77.1/87.5 & 65.3/87.5/54.7/69.2 & 43.2/22.4/37.3/34.3 \\

AA-CLIP~\cite{ma2025aa} & 71.3/60.4/85.1/72.3 & 82.7/70.1/91.7/81.5 & 83.7/83.4/85.4/84.2 & 92.8/93.4/86.6/90.9 & 79.3/86.9/63.8/76.7 & 53.3/16.6/51.1/40.3 \\
\midrule
\multicolumn{7}{c}{ \textbf{Unified Multi-scene, Multi-modal and Multi-class Model}} \\
\midrule
\rowcolor{blue!10}Ours &
96.4/95.9/96.8/96.3	&98.6/97.9/96.8/97.7	&93.0/94.5/95.0/94.2&	96.5/97.1/82.5/92.0	&85.4/92.3/48.0/75.2&	68.0/31.1/35.6/44.9 \\

\bottomrule
\end{tabular}

  \end{adjustbox}
\end{table*}

\begin{table*}[!htbp]
  \caption{	
  Per-Class comparison on the MulSen-AD~\cite{li2024multi} dataset for image-level metrics: $\text{AUC}_I $/$\text{AP}_I $ / $ \text{MF1}_I$. }
  \label{tab:mulsen_image}
  \centering
  \begin{adjustbox}{width=0.8\linewidth}
  \begin{tabular}{c|cccc|>{\columncolor{blue!10}}c}
\toprule
Method$\xrightarrow{}$ & MulSen-TripleAD~\cite{li2024multi} & AdaCLIP~\cite{cao2024adaclip} & MVFA~\cite{huang2024adapting} & AA-CLIP~\cite{ma2025aa} &  UniMMAD \\
Class$\downarrow$ & CVPR'25 &ECCV'24&CVPR'24&CVPR'25& Ours \\
\midrule
Cotton & \textbf{100.}/\textbf{100.}/\textbf{100.} & 47.3/81.1/87.6 & 54.3/83.5/87.6 & 47.3/81.1/87.6 & 99.7/99.6/98.7 \\

Cube & \textbf{100.}/\textbf{100.}/\textbf{100.} & 96.6/96.3/91.3 & 90.3/84.5/85.7 & 96.6/96.3/91.3 & 98.1/96.8/93.3 \\

Zipper & 99.1/99.6/97.2 & 69.8/88.5/82.3 & \textbf{100.}/\textbf{100.}/\textbf{100.} & 69.8/88.5/82.3 & \textbf{100.}/99.4/\textbf{100.} \\

Toothbrush & 91.3/96.3/89.8 & 34.0/37.6/66.7 & 44.4/49.4/65.4 & \textbf{98.4}/\textbf{99.0}/\textbf{96.2} & 81.4/80.5/85.2 \\

Spring pad & 67.5/86.2/85.7 & 74.4/70.1/76.9 & 91.0/91.0/84.8 & 85.8/92.9/\textbf{90.5} & \textbf{91.6}/\textbf{93.2}/86.6 \\

Piggy & \textbf{99.7}/\textbf{99.9}/\textbf{98.4} & 57.9/75.1/82.3 & 53.0/72.8/84.3 & 98.3/99.3/98.3 & 97.9/98.7/94.9 \\

Capsule & \textbf{96.8}/\textbf{99.3}/\textbf{96.0} & 52.4/83.8/89.1 & 74.9/91.5/88.5 & 87.0/97.2/91.4 & 94.2/98.1/95.8 \\

Light & 61.1/85.0/\textbf{89.2} & 38.0/51.7/74.0 & 49.7/54.3/75.0 & 45.2/79.9/87.8 & \textbf{80.1}/\textbf{86.8}/88.8 \\

Plastic cylinder & 99.6/\textbf{99.9}/98.2 & 88.3/94.6/88.9 & 71.7/85.1/80.8 & \textbf{100.}/99.5/\textbf{100.} & 99.3/99.1/97.9 \\

Screen & 20.0/60.5/\textbf{86.5} & 48.6/57.6/73.0 & \textbf{60.0}/55.6/78.6 & 15.3/\textbf{61.3}/86.4 & 45.3/60.7/79.8 \\

Screw & 92.9/97.3/96.8 & 56.2/49.3/53.8 & 65.9/51.0/62.2 & \textbf{100.}/\textbf{99.6}/\textbf{100.} & 75.0/60.6/82.4 \\

Flat pad & 95.3/\textbf{98.6}/93.5 & 84.6/86.8/88.9 & 51.3/64.0/76.7 & 95.6/98.4/\textbf{95.0} & \textbf{97.6}/97.9/93.8 \\

Nut & 33.4/71.8/85.3 & 43.2/29.7/53.1 & 61.5/42.6/61.9 & 46.3/\textbf{78.2}/\textbf{86.9} & \textbf{75.0}/72.9/79.9 \\

Button cell & 80.0/93.4/87.5 & 37.2/52.7/75.8 & 64.3/73.2/79.4 & \textbf{95.0}/\textbf{98.3}/\textbf{95.2} & 75.2/87.1/83.9 \\

Solar panel & 46.2/81.2/88.6 & 57.7/86.7/88.6 & 54.1/85.5/88.6 & \textbf{86.7}/\textbf{96.1}/\textbf{94.1} & 71.0/91.8/91.9 \\

\midrule
Mean & 78.8/\textbf{91.2}/\textbf{92.8} & 59.1/69.4/78.1 & 65.7/72.2/79.9 & 77.8/91.0/92.2 & \textbf{85.4}/88.2/90.2 \\

\bottomrule
\end{tabular}

  \end{adjustbox}
\end{table*}
\begin{table*}[!htbp]
   \caption{	
  Per-Class comparison on the MulSen-AD~\cite{li2024multi} dataset for  pixel-level metrics: $\text{AUC}_P $/$\text{MF1}_P $ / $ \text{AUPRO}$.  }
  \label{tab:mulsen_pixel}
  \centering
  \begin{adjustbox}{width=0.8\linewidth}
  \begin{tabular}{c|cccc|>{\columncolor{blue!10}}c}
\toprule
Method$\xrightarrow{}$ & MulSen-TripleAD~\cite{li2024multi} & AdaCLIP~\cite{cao2024adaclip} & MVFA~\cite{huang2024adapting} & AA-CLIP~\cite{ma2025aa} &  UniMMAD \\
Class$\downarrow$ & CVPR'25 &ECCV'24&CVPR'24&CVPR'25& Ours \\
\midrule
Cotton & 92.6/\textbf{30.8}/77.1 & 94.1/30.7/85.1 & 82.8/21.3/56.0 & 94.1/30.7/85.1 & \textbf{96.8}/25.6/\textbf{89.1} \\

Cube & \textbf{99.9}/\textbf{72.5}/\textbf{96.4} & 99.8/65.2/92.2 & 90.1/39.8/52.6 & 99.8/65.2/92.2 & 99.7/66.1/92.3 \\

Zipper & \textbf{99.1}/43.0/96.4 & \textbf{99.1}/\textbf{46.6}/\textbf{96.6} & 77.0/25.1/52.0 & \textbf{99.1}/\textbf{46.6}/\textbf{96.6} & 98.8/36.3/95.8 \\

Toothbrush & 99.2/\textbf{39.3}/94.8 & \textbf{99.3}/20.8/\textbf{95.9} & 63.9/\;\;4.6/36.3 & 98.2/14.1/91.5 & 98.2/27.0/89.2 \\

Spring pad & \textbf{99.7}/34.6/\textbf{98.0} & 99.0/25.9/96.6 & 98.3/\;\;3.3/94.2 & 93.1/34.1/84.3 & 94.8/\textbf{35.2}/88.0 \\

Piggy & 97.0/29.4/90.4 & 95.8/22.1/85.3 & 92.1/10.0/69.7 & 98.1/32.5/87.4 & \textbf{98.5}/\textbf{33.8}/\textbf{91.0} \\

Capsule & \textbf{99.3}/\textbf{30.0}/\textbf{97.0} & 99.1/25.6/\textbf{97.0} & 97.3/26.2/90.4 & 98.8/28.8/95.5 & 98.9/26.6/95.4 \\

Light & 96.7/16.9/91.7 & 97.3/20.0/93.9 & 95.8/\textbf{25.2}/76.8 & 97.2/19.5/92.7 & \textbf{98.5}/25.0/\textbf{94.7} \\

Plastic cylinder & \textbf{99.5}/47.4/\textbf{97.2} & 99.2/46.4/94.7 & 89.1/\;\;3.2/60.9 & 99.4/\textbf{50.3}/97.0 & 99.3/40.9/97.0 \\

Screen & 91.4/\;\;7.6/78.0 & \textbf{93.3}/11.2/\textbf{82.3} & 84.1/10.6/36.5 & 92.3/\textbf{14.1}/79.1 & 90.4/11.7/79.0 \\

Screw & 99.0/11.9/\textbf{96.5} & 98.6/11.2/95.6 & 97.3/16.8/88.8 & 98.9/12.7/\textbf{96.5} & \textbf{99.2}/\textbf{44.9}/96.1 \\

Flat pad & 99.6/\textbf{45.1}/97.7 & 99.2/35.9/97.0 & 97.8/\;\;6.4/93.0 & 99.4/43.4/97.4 & \textbf{99.7}/37.8/\textbf{98.2} \\

Nut & \textbf{99.6}/36.8/\textbf{97.1} & 98.9/16.7/96.0 & 98.0/\;\;2.2/94.0 & 98.5/34.1/92.4 & 98.8/\textbf{45.3}/89.6 \\

Button cell & \textbf{99.7}/38.9/98.0 & \textbf{99.7}/40.8/98.2 & 99.6/28.1/98.2 & 99.5/41.6/97.9 & \textbf{99.7}/\textbf{42.7}/\textbf{98.3} \\

Solar panel & 95.1/19.7/85.4 & 95.7/\textbf{25.6}/75.1 & 91.0/13.8/66.8 & 96.5/19.8/87.9 & \textbf{96.9}/20.5/\textbf{88.3} \\

\midrule
Mean & 97.8/33.6/\textbf{92.8} & 97.8/29.7/92.1 & 90.3/15.8/71.1 & 97.5/32.5/91.6 & \textbf{97.9}/\textbf{34.6}/92.1 
 \\

\bottomrule
\end{tabular}

  \end{adjustbox}
\end{table*}


\end{document}